\documentclass[sn-mathphys,Numbered]{sn-jnl}
\usepackage{graphicx}%
\usepackage{adjustbox} 
\usepackage{multirow}%
\usepackage{amsmath,amssymb,amsfonts}%
\usepackage{cleveref}
\usepackage{amsthm}%
\usepackage{mathrsfs}%
\usepackage[title]{appendix}%
\usepackage[ruled,linesnumbered]{algorithm2e}
\usepackage[table,x11names]{xcolor}
\usepackage{tabularx,ragged2e}
\newcolumntype{Y}{>{\centering\arraybackslash}X}
\usepackage[frozencache,cachedir=.]{minted}
\usepackage{tcolorbox}
\tcbuselibrary{minted,breakable,xparse,skins}

\definecolor{bg}{gray}{0.95}
\DeclareTCBListing{mintedbox}{O{}m!O{}}{%
  breakable=true,
  listing engine=minted,
  listing only,
  minted language=#2,
  minted style=default,
  minted options={%
    linenos,
    gobble=0,
    breaklines=true,
    breakafter=,,
    fontsize=\small,
    numbersep=8pt,
    #1},
  boxsep=0pt,
  left skip=0pt,
  right skip=0pt,
  left=25pt,
  right=0pt,
  top=3pt,
  bottom=3pt,
  arc=5pt,
  leftrule=1pt,
  rightrule=1pt,
  bottomrule=1pt,
  toprule=1pt,
  colback=bg,
  colframe=red!50,
  enhanced,
  overlay={%
    \begin{tcbclipinterior}
    \fill[red!15!white] (frame.south west) rectangle ([xshift=20pt]frame.north west);
    \end{tcbclipinterior}},
  #3}

\usepackage{textcomp}%
\usepackage{manyfoot}%
\usepackage{booktabs}%
\usepackage{algpseudocode}%
\usepackage{listings}%
\usepackage{siunitx}
\usepackage{subcaption}
\usepackage{stackengine}
\newmuskip\normalthickmuskip
\newmuskip\normalthinmuskip
\AtBeginDocument{%
  \normalthickmuskip=\thickmuskip
  \normalthinmuskip=\thinmuskip
}
\everymath{%
  \thickmuskip=\muexpr\normalthickmuskip-(1\normalthinmuskip plus 1\normalthinmuskip)\relax
}
\everydisplay{\thickmuskip=\normalthickmuskip}
\let\texdisplaystyle\displaystyle
\renewcommand{\displaystyle}{\texdisplaystyle\the\everydisplay}
\newcommand*\diff{\mathop{}\!\mathrm{d}}

\theoremstyle{thmstyleone}%
\theoremstyle{thmstyletwo}%

\theoremstyle{thmstylethree}%

\usepackage{xr-hyper}
\usepackage{hyperref}
\hypersetup{
    colorlinks=true,
    filecolor={blue},
    linkcolor=red,  
    citecolor=red,  
    urlcolor=blue,  
}
\usepackage{xr}
\let\cline\cmidrule
\makeatletter

\begin{document}
\emergencystretch 3em

\title[Article Title]{UniFIDES: Universal Fractional Integro-Differential Equation Solvers}


\author[1]{\fnm{Milad} \sur{Saadat}}\email{saadat.m@northeastern.edu}

\author[1]{\fnm{Deepak} \sur{Mangal}}\email{d.mangal@northeastern.edu}

\author*[1]{\fnm{Safa} \sur{Jamali}}\email{s.jamali@northeastern.edu}

\affil[1]{\orgdiv{Department of Mechanical and Industrial Engineering}, \orgname{Northeastern University}, \city{Boston}, \postcode{02115}, \state{Massachusetts}, \country{USA}}


\abstract{The development of data-driven approaches for solving differential equations has been followed by a plethora of applications in science and engineering across a multitude of disciplines and remains a central focus of active scientific inquiry. However, a large body of natural phenomena incorporates memory effects that are best described via fractional integro-differential equations (FIDEs), in which the integral or differential operators accept non-integer orders. Addressing the challenges posed by nonlinear FIDEs is a recognized difficulty, necessitating the application of generic methods with immediate practical relevance. This work introduces the Universal Fractional Integro-Differential Equation Solvers (UniFIDES), a comprehensive machine learning platform designed to expeditiously solve a variety of FIDEs in both forward and inverse directions, without the need for ad hoc manipulation of the equations. The effectiveness of UniFIDES is demonstrated through a collection of integer-order and fractional problems in science and engineering. Our results highlight UniFIDES' ability to accurately solve a wide spectrum of integro-differential equations and offer the prospect of using machine learning platforms universally for discovering and describing dynamical and complex systems.}

\keywords{Fractional integro-differential equations (FIDEs), Physics-informed machine learning, Fractional calculus, Neural networks}



\maketitle

\section{Introduction}\label{sec:Main}
Mathematical description of natural phenomena generally involves a series of differential equations (DEs). In the most generic form, partial (spatio-temporal) differential equations (PDEs) accept linear and nonlinear combinations of transient, flux, body force, and source terms, which are able to represent a wide array of dynamical systems. This is why development of algorithms that are geared towards improving general solutions of differential equations can have a significant impact across many fields. However, memory or spatially distributed quantities in general cannot be concisely described through DEs and are usually explained via integral terms. Integro-differential equations (IDEs), therefore, can describe systems with both non-local and pointwise effects.

IDEs have vital applications in electrical circuit analysis \cite{Barrios2019}, epidemics and population balances \cite{Glasgow2014,Bonnefon2014}, renewable energy \cite{Sidorov2020}, potential theory \cite{Wazwaz2011}, and thermo-fluid sciences \cite{MacCamy1977,AzZobi2021}, among others. Many crucial tasks in signal processing \cite{Kumar2017}, material design \cite{Li2017a,Ferras2018,Dabiri2023}, and economics \cite{Kilbas2006,Traore2020}, nonetheless, are described through \textit{fractional} integro-differential equations (FIDEs), in which the integral or derivative operators accept any arbitrary real or complex order \cite{Sun2018a}. Therefore, FIDEs can not only arguably capture the broadest and most complex set of phenomena but also often offer the most mathematically concise/compact descriptions. However, solving FIDEs has proven to be far from trivial, and algorithms that can directly solve FIDEs can be transformative in a foundational way.

Methods to solve IDEs can be categorized into analytical, semi-analytical, and numerical techniques. Analytical methods, such as Laplace transforms \cite{Thorwe2012} and Taylor series \cite{Huang2011}, often yield precise, closed-form solutions but are limited to relatively simpler cases \cite{Sunthrayuth2021}. Semi-analytical techniques, including Adomian decomposition \cite{Hashim2006}, homotopy perturbation \cite{Saberi2009,Das2019}, and variational iteration methods \cite{Wang2007a}, can tackle a broader range of linear and nonlinear problems, though they may present challenges in problem formulation and convergence. Numerical techniques, such as Chebyshev polynomials \cite{Babaei2020,Atta2022}, Haar wavelet \cite{Aziz2013}, Galerkin \cite{Chen2020}, and finite difference methods \cite{Abbaszadeh2021}, are highly versatile and adaptable for a wide array of problems. However, these methods can be complex to implement, approximate the solution, and are often computationally intensive.

In recent years, machine learning (ML) tools have shown great potential in addressing complex computational problems across various domains \cite{Jiang2023,Meuris2023,Thiyagalingam2022}. By incorporating prior knowledge, such as principled physical laws that dictate the spatio-temporal dynamics of systems, science-aware machine learning is extensively being applied to a wide set of challenges in science and engineering \cite{Karniadakis2021}. For instance, by leveraging automatic differentiation (AD) to directly solve DEs, Physics-Informed Neural Networks (PINNs) were developed as robust platforms for data-driven science and engineering computations \cite{Raissi2019,Raissi2020}. The discovery of governing equations of a system from data was also shown to be feasible using sparse identification of nonlinear dynamics \cite{Kaptanoglu2022}. Moreover, mesh-based discretization methods have proven effective in solving forward and inverse PDE problems \cite{Karnakov2022}. Neural operators and transformers are also gaining traction for their ability to solve a family of PDEs instead of a single PDE, reducing the overall computation time \cite{Li2022,Li2020,Lu2021a}.

Despite the successful endeavors in developing ML frameworks for the solution of differential equations, there appears to be a general scarcity of equivalent platforms for addressing integral equations. A few ML tools have been employed to solve IDEs \cite{Fu2022,Lu2021} and fractional DEs \cite{Sivalingam2023}. In these examples, the integral (or fractional derivative) operators are approximated using a numerical discretization scheme such as Simpson's rule or Gaussian quadrature. In \cite{Pang2019}, fractional-PINN (fPINN) was introduced and used to solve fractional DEs in forward and inverse directions. In a recent PINN attempt to solve a wider set of IDEs, the auxiliary-PINN (A-PINN) method was introduced \cite{Yuan2022}, which defines an auxiliary equation to represent the integral term. This auxiliary equation is then converted to an ordinary differential equation (ODE) using AD of the auxiliary output. By employing the A-PINN method, the need for integral approximation is cleverly obviated, demonstrating promise in solving a variety of IDEs. However, A-PINN is restricted to integer-order IDEs, as fractional differentiation is not feasible with AD; see \Cref{sec:Methods}. Fractional integrals and differentiations, therefore, inevitably demand some form of numerical approximation.

This study enhances the effectiveness of physics-informed neural networks (PINNs) by introducing a Universal Fractional Integro-Differential Equation Solver, UniFIDES. For the first time, a versatile PINN-based framework is introduced, adept at solving nonlinear and multi-dimensional integer-order and fractional integro-differential equations (FIDEs), including systems of FIDEs and partial FIDEs. Streamlined with automatic differentiation for integer-order differentiation, UniFIDES can solve both forward and inverse problems. Through extensive experiments on various integer-order and fractional Fredholm and Volterra integral and integro-differential equations, UniFIDES demonstrates its robustness and precision. By eliminating the need for any mathematical manipulation, UniFIDES offers out-of-the-box functionality for a diverse set of problems in science and engineering.

\section{Results}\label{sec:Results}
Unlike other FIDE frameworks that usually demand ad-hoc modifications for each type of problem, UniFIDES operates on a seamless plug-and-play basis; the equations are embedded as-is, and the imposition of boundary and initial conditions is intuitive and straightforward. To manifest the generality of UniFIDES, we begin by recalling the fractional integral of order $\alpha \in \mathbb{R^+}$ in the Riemann-Liouville (RL) sense\footnote{In fact, $\alpha$ can be complex, but for most engineering and scientific applications, $\alpha$ remains real.} \cite{Diethelm2005}:

\begin{equation}
    [{}^\alpha \mathcal{I}_0^x]{u(x)}=\frac{1}{\Gamma{(\alpha)}}\int_0^x \frac{1}{(x-t)^{1-\alpha}}u(t)\diff t
    \label{eq:RLI}
\end{equation}
where $x \in \mathbb{R^+}$, $\Gamma(\cdot)$ is Euler's continuous gamma function, and $u(t)$ is a non-singular function in general. \Cref{eq:RLI} is essentially a fractional Volterra IE of order $\alpha$ whose kernel has a strong singularity at one endpoint of the integration interval. For $\alpha=1$, \Cref{eq:RLI} reduces to a typical integral expression. The RL fractional derivative of order $\beta$ w.r.t. $x$, $[{}^\beta \mathcal{D}_x]$, will then be defined as follows \cite{Baleanu2016}:

\begin{equation}
    [{}^\beta \mathcal{D}_x]{u(x)}=\frac{\diff^m}{\diff x^m}[{}^{m-\beta}\mathcal{I}] u(x)=\frac{1}{\Gamma{(-\beta)}}\int_0^x \frac{1}{(x-t)^{1+\beta}}u(t)\diff t = [{}^{-\beta} \mathcal{I}_0^x]{u(x)}
    \label{eq:RLD}
\end{equation}
where $m=\lceil \beta \rceil$ denotes the ceiling function.\footnote{The fractional derivative in the Caputo sense is also accessible by swapping the sequence of differentiation and integration in the first equality in \Cref{eq:RLD}, which immediately yields a corresponding definition in the RL sense and vice versa; see the second chapter in \cite{Baleanu2016}.} The second equality in \Cref{eq:RLD} is valid for suitable $u(x)$ functions; see \cite{Baleanu2016}. In fact, fractional derivative of order $\beta$ equivalently corresponds to a fractional integral of order $-\beta$; compare the last equality in \Cref{eq:RLD} with \Cref{eq:RLI}.

\begin{figure}
  \centering
  \includegraphics[width=.9\linewidth]{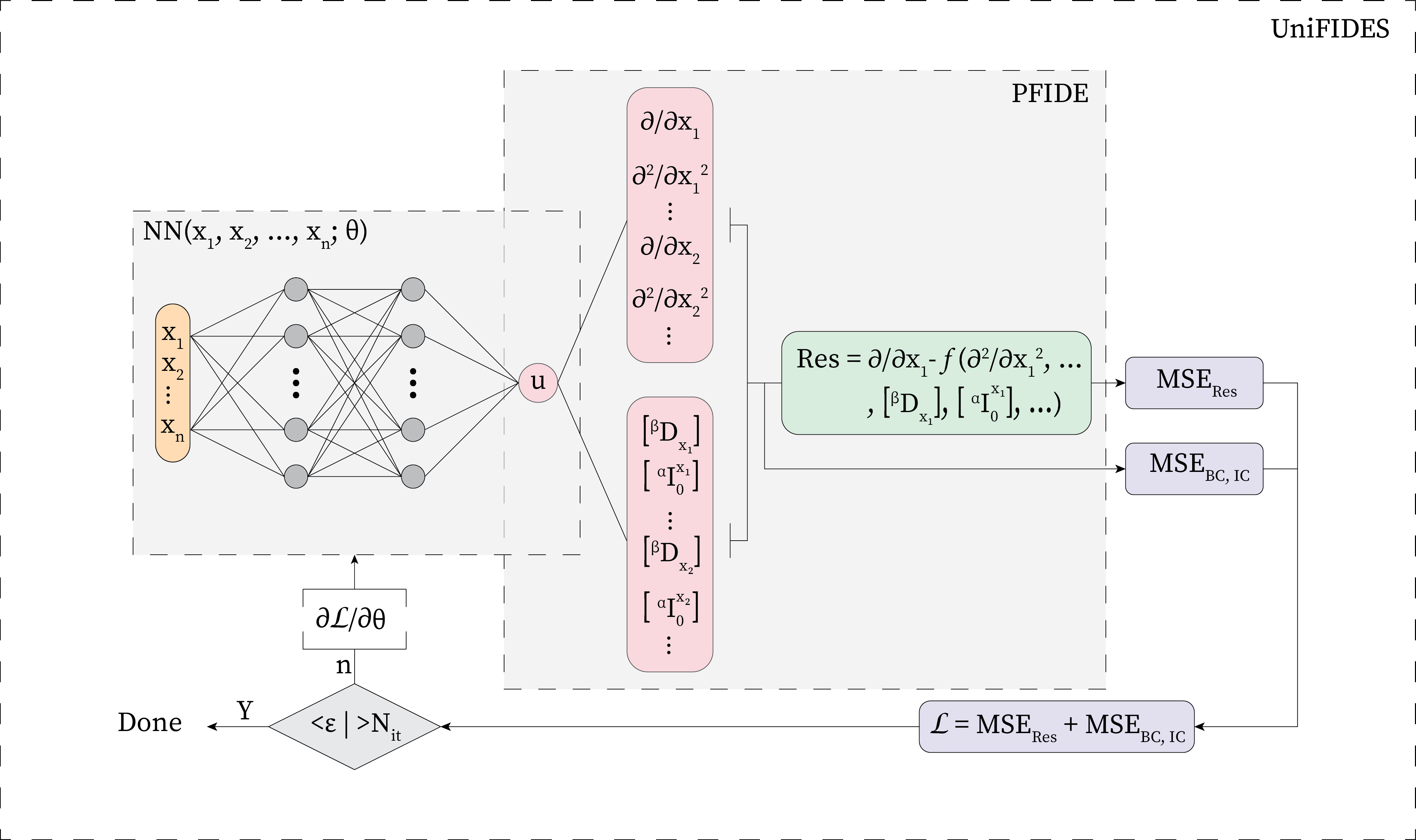}
  \caption{The general architecture of UniFIDES for solving the forward problem of partial fractional integro-differential equations (PFIDEs). The same architecture is slightly modified to solve inverse problems. The integer-order derivatives are handled using AD (the top narrow box), while integrals and fractional derivatives are calculated through the numerical scheme introduced in \Cref{sec:Methods}. The training is halted once the relative error plateaus or if the maximum number of iterations ($N_\mathrm{it}$) is reached. The training process and hyperparameters are described in detail in \Cref{sec:SI-train}.}
  \label{fig:Arch}
\end{figure}

In summary, \Cref{eq:RLI} serves as the fundamental expressions to calculate integer-order and fractional integrals ($\alpha\geq 0$), while \Cref{eq:RLD} handles the fractional derivatives ($0\leq\beta<1$) in FIDEs. For integer-order derivatives, AD is adopted. In \Cref{sec:Methods}, the numerical methods to approximate these fractional operators are put forth. The architecture of UniFIDES is shown in \Cref{fig:Arch}, with more information about the training process and hyperparameters in \Cref{sec:SI-train}.

In the results that follow, four forward integer-order IE and IDE problems are introduced. Then, three more involved fractional IE, IDE, and system of IDEs are solved in the forward direction. Finally, the inverse solution of the system of FIDEs is presented as the most complicated of all test cases, with the objective of recovering the fractional operator orders. Additional cases can be found in \ref{sec:SI-Res}.

\subsection*{Integer-order forward solutions}
\label{sec:Res_Int}
\subsubsection*{Forward solution of 1D Fredholm IDE}
\label{sec:Res_F1}
\begin{figure}
  \centering
  \begin{subfigure}[t]{0.3\textwidth}
    \stackinset{l}{0.2in}{t}{-.35in}{(a)}%
    {\includegraphics[width=\textwidth, height=3.5cm, keepaspectratio]{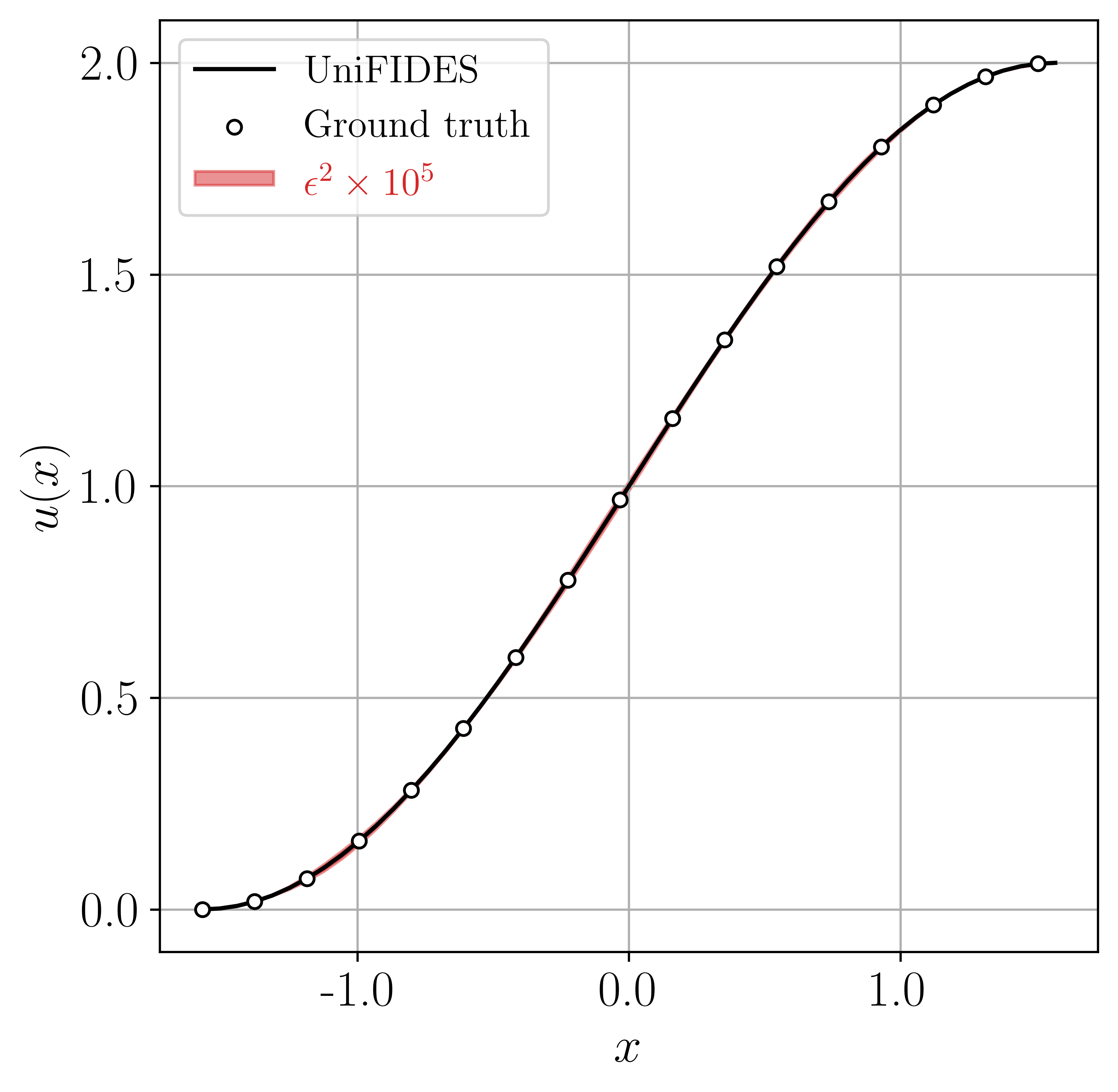}} 
    \phantomsubcaption
    \label{fig:Res_C1_C2_a}
  \end{subfigure}\hfill 
  \begin{subfigure}[t]{0.7\textwidth}
    \stackinset{l}{0.2in}{t}{-.05in}{(b)}%
    {\includegraphics[width=\textwidth, height=5cm, keepaspectratio]{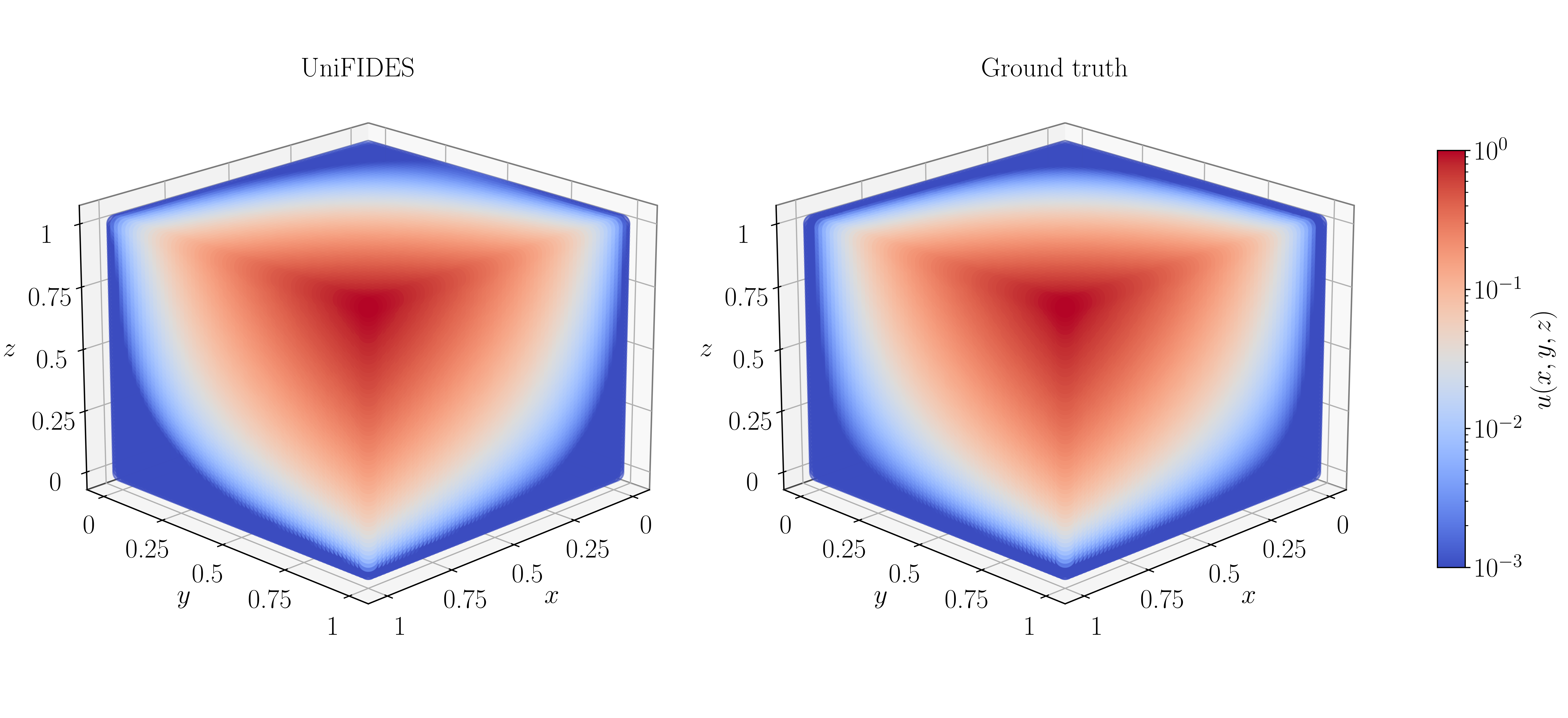}} 
    \phantomsubcaption
    \label{fig:Res_C1_C2_b}
  \end{subfigure}
  \caption{The solution by UniFIDES and the exact solution for (a) 1D Fredholm IDE (Case 1) and (b) 3D Fredholm IE (Case 2). The absolute squared error ($\epsilon^2$) for line graphs is shown with red shades and is multiplied by a constant in this work (\num{1e5} in this case). In panel (b), the prediction MSE is \num{1.07e-6}.
}
  \label{fig:Res_C1_C2}
\end{figure}

Case 1 is that of reference \cite{Yuan2022}, which is an integer-order nonlinear 1D Fredholm IDE with applications in diffusion processes and quantum mechanics:

\begin{equation}
\left\{
\begin{aligned}
&\left[ {}^{1} \mathcal{D}_x \right] u(x) = \cos{x} - x + \frac{1}{4}\left[ {}^1 \mathcal{I}_{-1/4}^{1/4} \right] xt u^2(t) \, \mathrm{d}t \\
&x \in \left[ -\frac{\pi}{2}, \frac{\pi}{2} \right]\\
&u\left( -\frac{\pi}{2} \right)=0
\end{aligned}
\right.
\label{eq:F1}
\end{equation}
The exact solution reads $u(x) = 1 + \sin{x}$. This is a forward problem, and the objective is to find $u(x)$. \Cref{eq:F1} is directly implemented in its continuous form with the same PINN hyperparameters and collocation points as in \cite{Yuan2022}; see \Cref{sec:SI-train} for a summary of the hyperparameters. The integer-order derivative is handled using AD, while the integral term is approximated through the numerical scheme introduced in \Cref{eq:RL,eq:RL_c}. The UniFIDES prediction is shown in \Cref{fig:Res_C1_C2_a}. Despite the inevitable truncation error associated with the numerical approximation of the integral term, UniFIDES still managed to yield an accurate prediction compared to A-PINN: The relative L2 error for this case was reported to be \SI{0.048}{\percent} using A-PINN, while this value is \SI{0.019}{\percent} with UniFIDES. The mean squared errors (MSEs) for all cases tested are summarized in \Cref{tab:MSE} and benchmarked against integer-order cases using A-PINN; see \Cref{sec:SI-APINN} for more details on the A-PINN training procedure and potential limitations of it.

{\renewcommand{\arraystretch}{1.2}
\begin{table}[h]
\centering
\caption{The summary of cases tested in this work. The order doublets $(\alpha, \beta)$ correspond to integral and differentiation orders, respectively. The number of residual points ($N$) is also reported. The integer-order cases are also implemented using A-PINN with the same set of hyperparameters and benchmarked against each other, with the method that yields the lower MSE being highlighted. see \Cref{sec:SI-train,sec:SI-APINN} for more details.}
\label{tab:MSE}
\begin{tabular}{clcclccc}
\hline\hline
Case & Type & $(\alpha,\beta)$ & Dim. & $N$ (Eq. \ref{eq:RL}) & \multicolumn{2}{c}{MSE} & Ref. \\
\hline
     &      &                  &      &                        & UniFIDES & A-PINN &      \\\cline{6-7}
1    & Fredholm IDE     & (1, 1)    & 1D    & 50 & \num{5.20e-8} & \cellcolor{gray!20}\num{8.84e-10} & \cite{Yuan2022}    \\
2    & Fredholm IE     & (1, --)    & 3D    & $10\times 10\times 10$ & \cellcolor{gray!20}\num{1.07e-6} & \num{2.45e-5} & \cite{Mahdy2023}   \\
3    & Volterra IDE     & (1, 1)    & 1D    & 64 & \cellcolor{gray!20}\num{8.78e-6} & \num{1.23e-4} & \cite{Feldstein1974}    \\
4    & Volterra IE     & (1, --)    & 2D    & $5\times 8$ & \num{1.50e-5} & \cellcolor{gray!20}\num{3.18e-6} & \cite{Saberi2011}    \\
\hline
5    & Volterra FIE     & (0.5, --)    & 1D    & 64 & \cellcolor{gray!20}\num{1.13e-6} & NA & \cite{Usta2021}    \\
6    & Volterra FIDE     & (1, 0.7)    & 2D    & $8\times 8$ & \cellcolor{gray!20}\num{1.17e-3} & NA & \cite{Santra2022}    \\
7    & Sys. of Volterra FIDEs     & (1, 0.5)    & 1D & 64 & \cellcolor{gray!20}\num{7.55e-7} & NA & \cite{Akbar2020}    \\
\hline\hline
\end{tabular}
\end{table}

\subsubsection*{Forward solution of 3D Fredholm IE}
\label{sec:Res_F2}
To further demonstrate UniFIDES' generality to higher-dimensional problems, Case 2 is deemed to be a 3D Fredholm IE reported by reference \cite{Mahdy2023}:

\begin{equation}
\left\{
\begin{aligned}
&\begin{aligned}
    u(x,y,z) &= x^2y^2z^2 - \frac{1}{\num{29400}}e^{-xyz} \\
    & + 0.01\left[ {}^1 \mathcal{I}_{0}^{1} \right]\left[ {}^1 \mathcal{I}_{0}^{1} \right]\left[ {}^1 \mathcal{I}_{0}^{1} \right]e^{-xyz}t^2sr^2u^2(t,s,r)\diff t\diff s\diff r
\end{aligned} \\
&(x,y,z) \in \left[0, 1\right]\\
&u\left(0,0,0\right)=0
\end{aligned}
\right.
\label{eq:F2}
\end{equation}
with an exact solution of $u(x,y,z)=x^2y^2z^2$. Such IEs have applications in electromagnetic theory and non-homogeneous elasticity. As can be seen in \Cref{fig:Res_C1_C2_b}, the UniFIDES' prediction closely mimics the exact solution with an excellent MSE of \num{1.07e-6}. As an additional case, the same IE but for $(x,y,z) \in [0,2]$ is solved, and despite the accuracy deterioration often reported when dealing with irregular ranges for PINN-based frameworks, UniFIDES still managed to recover the 3D dynamics with an MSE of \num{7.85e-3}; see \Cref{sec:SI-Res}.

\subsubsection*{Forward solution of 1D Volterra IDE}
\label{sec:Res_V3}
Fredholm IDEs, with fixed integration limits, reflect aggregate system properties, while Volterra IDEs, with a variable upper limit, take into account the system's immediate history up to the current point. This makes Volterra IDEs ideal for applications in population dynamics and epidemic modeling \cite{Wazwaz2011} and also differentiates their integral modeling; see \Cref{sec:SI-FV}. For instance, the following nonlinear 1D Volterra IDE (Case 3) \cite{Feldstein1974} has applications in population growth of species and financial mathematics:

\begin{equation}
\left\{
\begin{aligned}
&\left[ {}^{1} \mathcal{D}_x \right] u(x) = \frac{5}{2}x - \frac{1}{2}xe^{x^2} + \left[ {}^1 \mathcal{I}_{0}^{x} \right] xte^{u(t)}\diff t \\
&x \in \left[ 0, 1 \right]\\
&u\left(0\right)=0
\end{aligned}
\right.
\label{eq:V3}
\end{equation}
with an exact solution of $u(x)=x^2$. For each point $x$ between 0 and 1, the integral accumulates the effect of the integrand $xte^{u(t)}$ over the interval from $t=0$ to $x$. This indicates that what happens between $t=0$ and $x$ has a cumulative impact on the outcome at $x$, meaning that the integral term represents non-local interactions within the system. Again, the integer-order derivative is obtained using AD. The UniFIDES solution of this IDE is shown in \Cref{fig:Res_C3_C4_a}, with an MSE of \num{8.78e-6}. Due to the hereditary nature of fractional operators, the squared error increases monotonically along the $x$ direction. To ameliorate this inevitable artifact, the number of residual points can be increased, which corresponds to a smaller value of step size, $h$; see \Cref{sec:Methods}.

\begin{figure}
  \centering
  \begin{subfigure}[t]{0.305\textwidth}
  \stackinset{l}{0.2in}{t}{-.2in}{(a)}
  {\includegraphics[width=\textwidth]{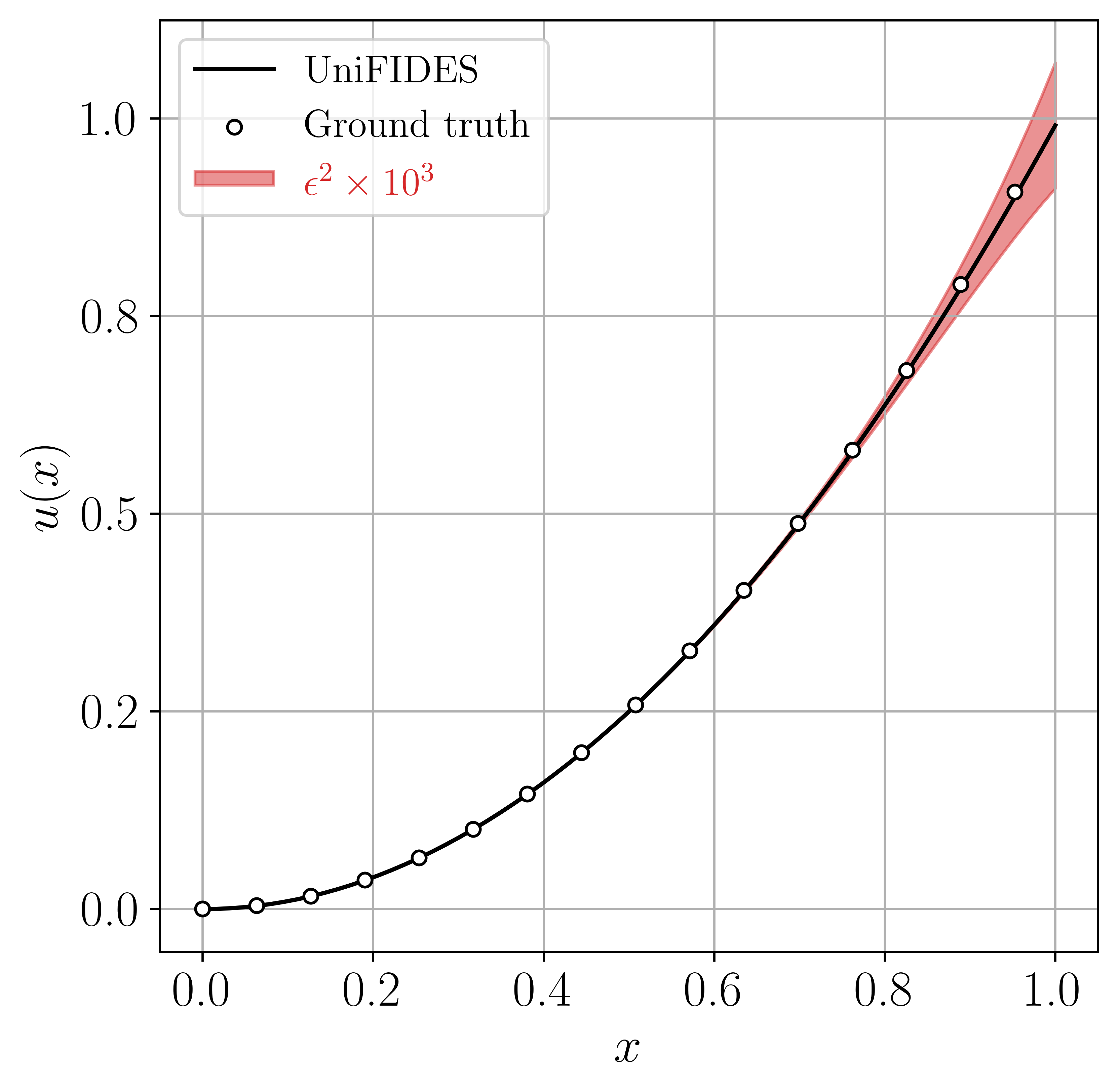}}
    \phantomsubcaption
  \label{fig:Res_C3_C4_a}
  \end{subfigure}\hfill
  \begin{subfigure}[t]{0.67\textwidth}
  \stackinset{l}{0.2in}{t}{-.13in}{(b)}
  {\includegraphics[width=\textwidth]{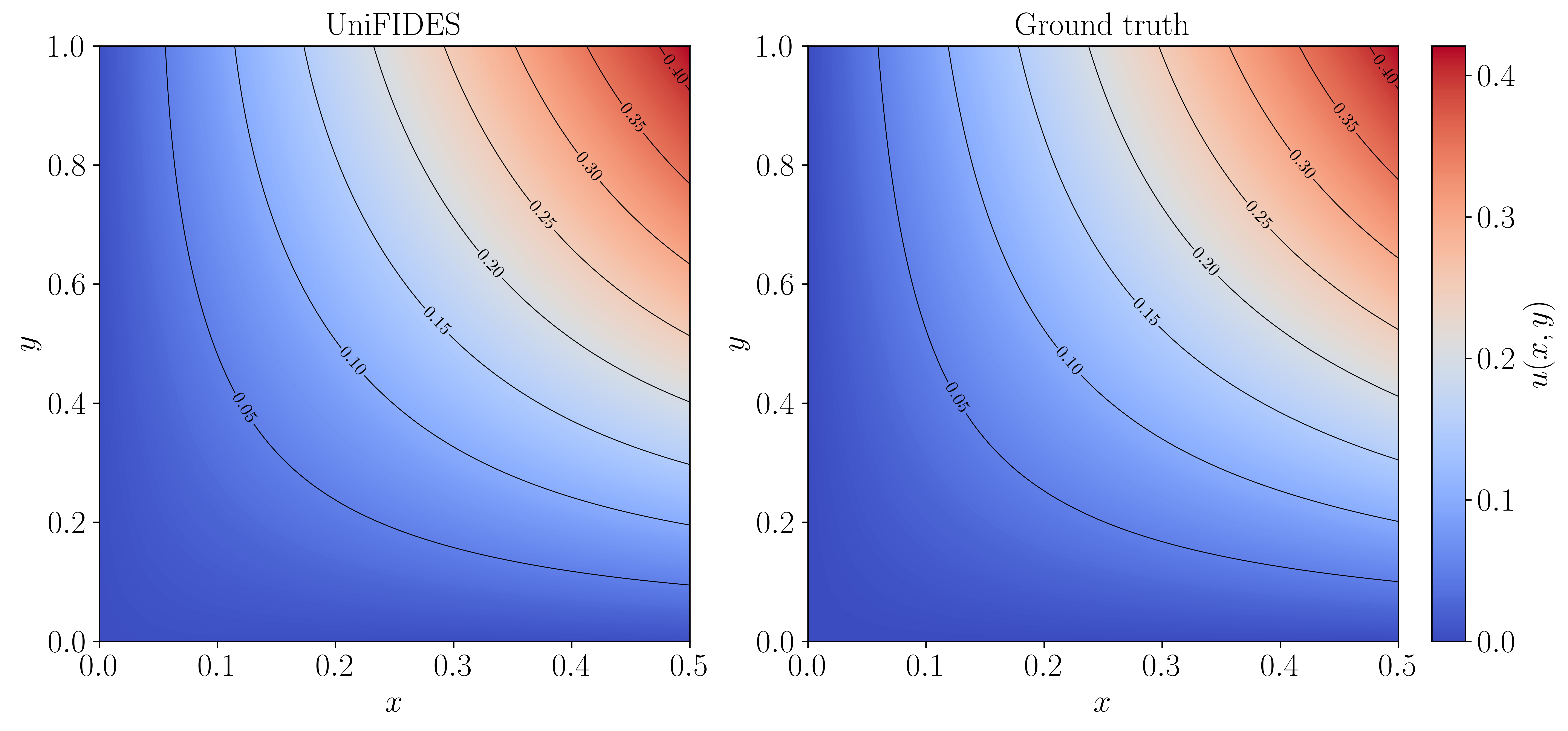}}
    \phantomsubcaption
  \label{fig:Res_C3_C4_b}
  \end{subfigure}
  \caption{The solution by UniFIDES and the exact solution for (a) 1D Volterra IDE (Case 3) and (b) 2D Volterra IE (Case 4).}
  \label{fig:Res_C3_C4}
\end{figure}

\subsubsection*{Forward solution of 2D Volterra IE}
\label{sec:Res_V4}
The following 2D nonlinear Volterra IE is utilized as Case 4 to assess the applicability of UniFIDES to higher-dimension Volterra cases \cite{Saberi2011}:

\begin{equation}
\left\{
\begin{aligned}
&u(x,y)= f(x,y) + \left[ {}^1 \mathcal{I}_{0}^{y} \right]\left[ {}^1 \mathcal{I}_{0}^{x} \right] \left(xt^2+\cos(s)\right)u^2(t,s) \diff t \diff s \\
&f(x,y)=x\sin(y)\left(1-\frac{x^2\sin^2{y}}{9}\right)+\frac{x^6}{10}\left(\frac{\sin(2y)}{2}-y\right)\\
&(x, y) \in \left[ 0, 0.5 \right], \left[ 0, 1 \right]\\
&u\left(0, 0\right)=0
\end{aligned}
\right.
\label{eq:V4}
\end{equation}
Such IEs are seen in plane contact of inhomogeneous, ageing viscoelastic materials. The exact solution reads $u(x,y)=x\sin y$. The $x$ and $y$ upper limits are chosen to be \num{0.5} and \num{1}, respectively, to demonstrate UniFIDES' versatility in handling double integrals with different step sizes. The UniFIDES solution, along with the exact solution, are presented in \Cref{fig:Res_C3_C4_b}. The solution attained an MSE of \num{1.50e-5}.

\subsection*{Fractional-order forward and inverse solutions}
\label{subsec:Res_Frac}
So far, the test cases have only included integer-order differentiation and integral operators. The derivatives were handled by AD, while the integrals were calculated using the numerical approximation explained in \Cref{sec:Methods}. In the following test cases, the same numerical scheme but for fractional-order operators is utilized. Despite the increasing complexity of FIDEs compared to IDEs, UniFIDES is agnostic to the operator orders and nonlinearity. Additionally, attention is given exclusively to fractional Volterra instances due to their heightened intricacy.

\subsubsection*{Forward solution of 1D Volterra FIE}
\label{sec:Res_V5}
The first fractional problem (Case 5) is a 1D Volterra FIE \cite{Usta2021}:

\begin{equation}
\left\{
\begin{aligned}
&u(x) = \sqrt{\pi}\left(1+x\right)^{-1.5} - 0.02\frac{x^3}{1+x} + 0.01x^{2.5}\left[ {}^{0.5} \mathcal{I}_{0}^{x} \right] u(t) \diff t\\
&x \in \left[ 0, 4 \right]\\
&u\left(0\right)=\sqrt{\pi}
\end{aligned}
\right.
\label{eq:V5}
\end{equation}
The integral operator in this case has a fractional order ($\alpha = 0.5$), and the exact solution reads $u(x) = \sqrt{\pi}(1 + x)^{-1.5}$. Such FIEs are frequently seen in crystal growth and heat conduction. Here, the range of $x$ is also extended. As shown in \Cref{fig:Res_C5_C6_a}, the UniFIDES prediction closely mimics the reference, with an MSE of \num{1.13e-6}. Again, error accumulation is witnessed as $x$ increases, which can be mitigated by reducing the step size, $h$; see \Cref{sec:Methods}. The error magnitude nonetheless is negligible for most engineering and scientific applications.

\begin{figure}
  \centering
  \begin{subfigure}[t]{0.305\textwidth}
  \stackinset{l}{0.2in}{t}{-.2in}{(a)}
  {\includegraphics[width=\textwidth]{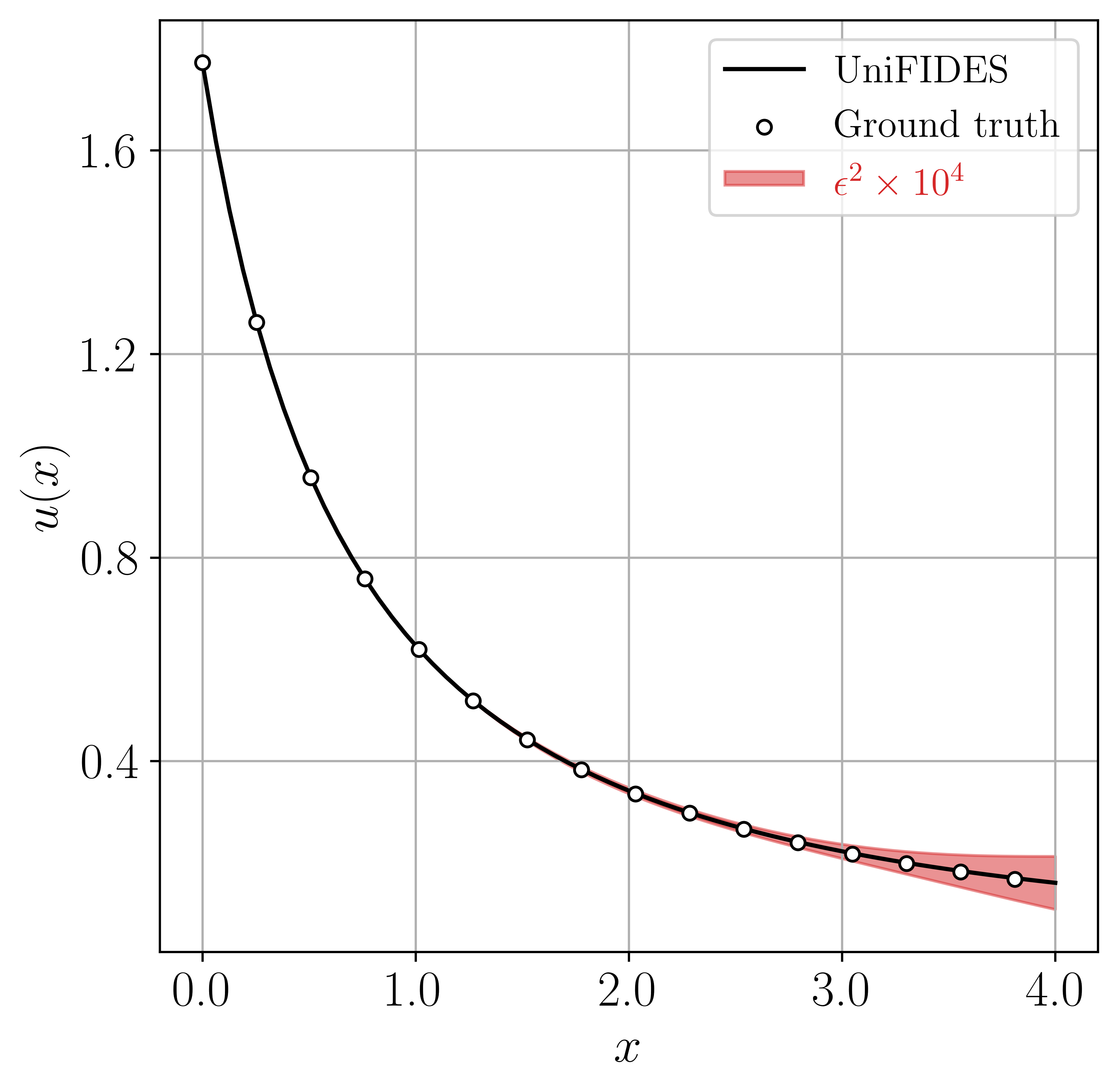}}
    \phantomsubcaption
  \label{fig:Res_C5_C6_a}
  \end{subfigure}\hfill
  \begin{subfigure}[t]{0.67\textwidth}
  \stackinset{l}{0.2in}{t}{-.13in}{(b)}
  {\includegraphics[width=\textwidth]{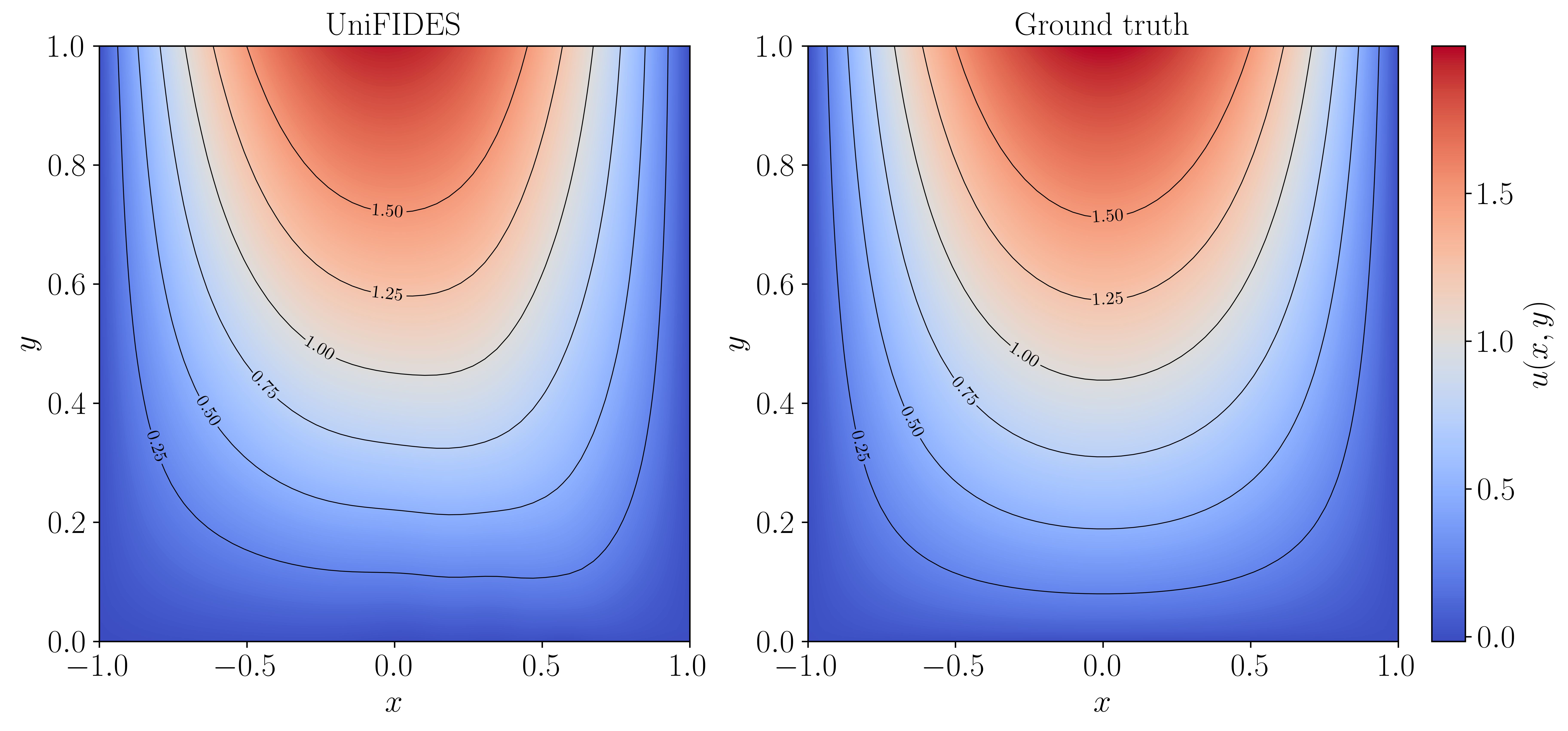}}
    \phantomsubcaption
  \label{fig:Res_C5_C6_b}
  \end{subfigure}
  \caption{The solution by UniFIDES and the exact solution for (a) 1D Volterra FIDE (Case 5) and (b) 2D Volterra partial FIDE (Case 6) for $\beta=0.7$.}
  \label{fig:Res_C5_C6}
\end{figure}

\subsubsection*{Forward solution of 2D Volterra partial FIDE}
\label{sec:Res_V6}
Case 6 is a nonlinear 2D Volterra FIDE which involves derivatives of both inputs, therefore covering \textit{partial} FIDEs \cite{Santra2022}:

\begin{equation}
\left\{
\begin{aligned}
&\left[ {}^{\beta} \mathcal{D}_y \right] u(x,y) - \frac{\partial^2 u}{\partial x^2} +\left[ {}^1 \mathcal{I}_{0}^{y} \right] x(y-s)u(x,s) \diff s = f(x,y)\\
&\begin{aligned}
    f(x,y)&=\left(1-x^2\right)\left(\frac{y^{1-\beta}}{\Gamma(2-\beta)}+\Gamma(1+\beta)\right) \\
    & + 2\left(y+y^\beta\right)+x\left(1-x^2\right)\left(\frac{y^3}{6}+\frac{y^{2+\beta}}{(1+\beta)(2+\beta)}\right)
\end{aligned} \\
&(x, y) \in \left[ -1, 1 \right], \left[ 0, 1 \right]\\
&u\left(-1, y\right)=u\left(1, y\right)=0\\
&u\left(x, 0\right)=0
\end{aligned}
\right.
\label{eq:V6}
\end{equation}
The exact solution is $u(x,y)=(1-x^2)(y-y^\beta)$. Such partial FIDEs are encountered in grain growth and reactor dynamics.\footnote{By replacing $y$ with $t$, this partial FIDE can be thought of as a \textit{time-fractional} IDE. However, $t$, $s$, and $r$ are reserved for dummy integral variables in this work.} The UniFIDES solution for $\beta=0.7$ is presented in \Cref{fig:Res_C5_C6_b}. Despite the increased complexity of this case, UniFIDES recovered the exact solution with an MSE of \num{1.17e-3}, which reduces for finer grids. As before, the second-order derivative w.r.t. $x$ is handled using AD, while the integral and fractional derivative terms are approximated using the numerical scheme. Here, attention should be given once partial fractional derivatives are calculated to ensure numerical accuracy; see \Cref{sec:SI-Deriv}.

\subsubsection*{Forward solution of a system of Volterra FIDEs}
\label{sec:Res_V7}
As the final forward solution problem, the objective in Case 7 is to solve a system of nonlinear Volterra FIDEs:

\begin{equation}
\left\{
\begin{aligned}
&\begin{aligned}
    \left[ {}^{\beta} \mathcal{D}_x \right]u_1(x) & - \frac{3x^{2\beta}\beta\Gamma(3\beta)}{\Gamma(1+2\beta)}\\
    &-\left[ {}^1 \mathcal{I}_{0}^{x} \right]\left(x-t\right)u_1(t) \diff t -\left[ {}^1 \mathcal{I}_{0}^{x} \right]\left(x-t\right)u_2(t) \diff t = 0\\
\end{aligned} \\
&\begin{aligned}
    \left[ {}^{\beta} \mathcal{D}_x \right]u_2(x) & +\frac{2x^{2+3\beta}}{2+9\beta+9\beta^2}+\frac{3x^{2\beta}\beta\Gamma(3\beta)}{\Gamma(1+2\beta)}\\
    &-\left[ {}^1 \mathcal{I}_{0}^{x} \right]\left(x-t\right)u_1(t) \diff t -\left[ {}^1 \mathcal{I}_{0}^{x} \right]\left(x-t\right)u_2(t) \diff t = 0\\
\end{aligned} \\
&x \in \left[0, 1\right]\\
&u_1\left(0\right)=u_2\left(0\right)=0
\end{aligned}
\right.
\label{eq:V7}
\end{equation}
with an exact solution of $u_1(x)=x^{3\beta}$ and $u_2(x)=-x^{3\beta}$. Such coupled FIDEs arise in enzyme kinetics and gas-liquid reaction problems. The UniFIDES solution for $\beta=0.5$ is presented in \Cref{fig:Res_C7}. The MSE for this case was \num{7.55e-7}, which demonstrates UniFIDES' generality for multi-output cases.

\begin{figure}
    \centering
    \begin{subfigure}[b]{0.47\textwidth}
        \stackinset{l}{0.2in}{t}{-.2in}{(a)}{\includegraphics[width=\textwidth]{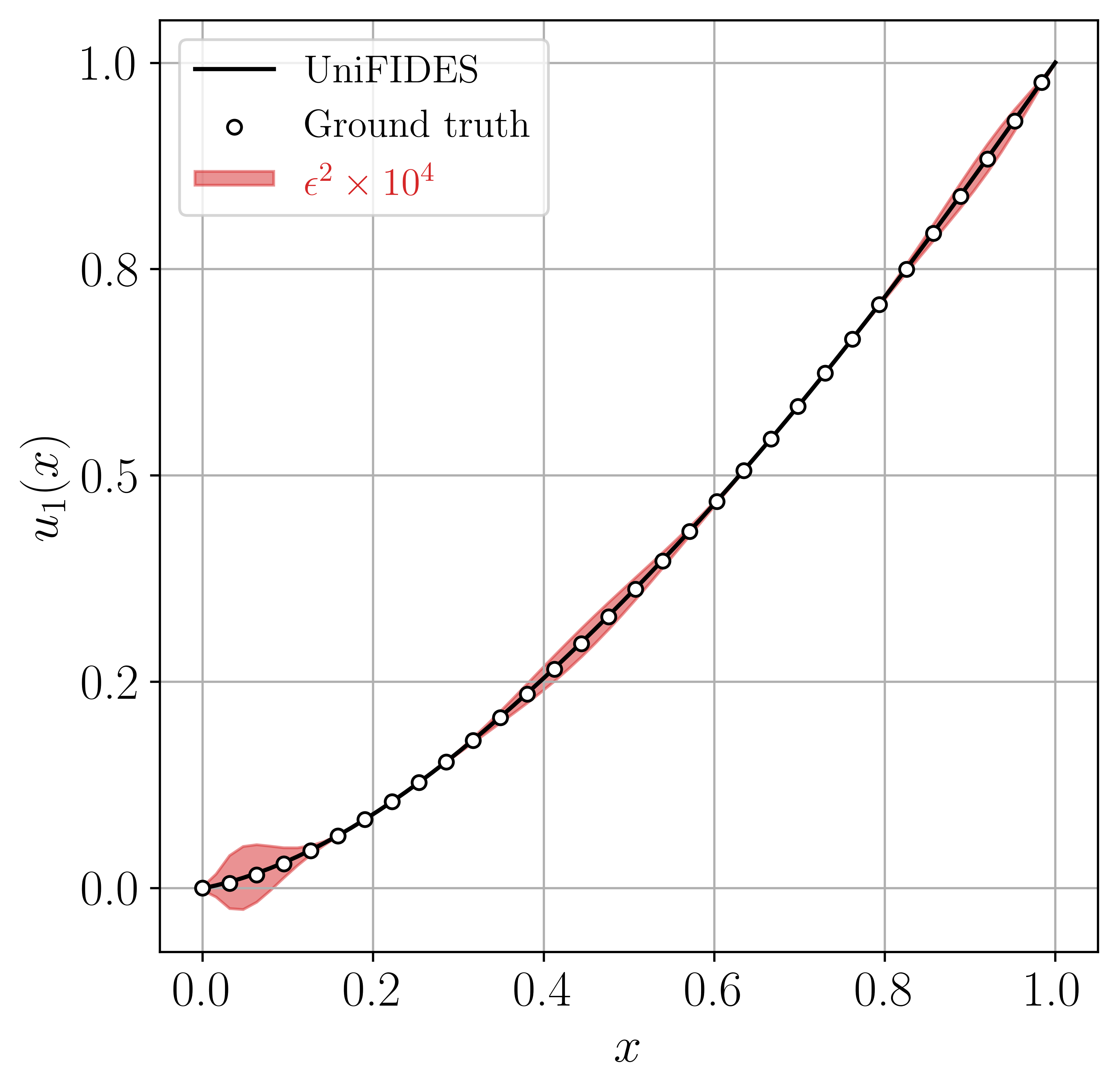}}
        \phantomsubcaption
        \label{fig:Res_C7_a}
    \end{subfigure}
    \hfill 
    \begin{subfigure}[b]{0.47\textwidth}
        \stackinset{l}{0.2in}{t}{-.2in}{(b)}{\includegraphics[width=\textwidth]{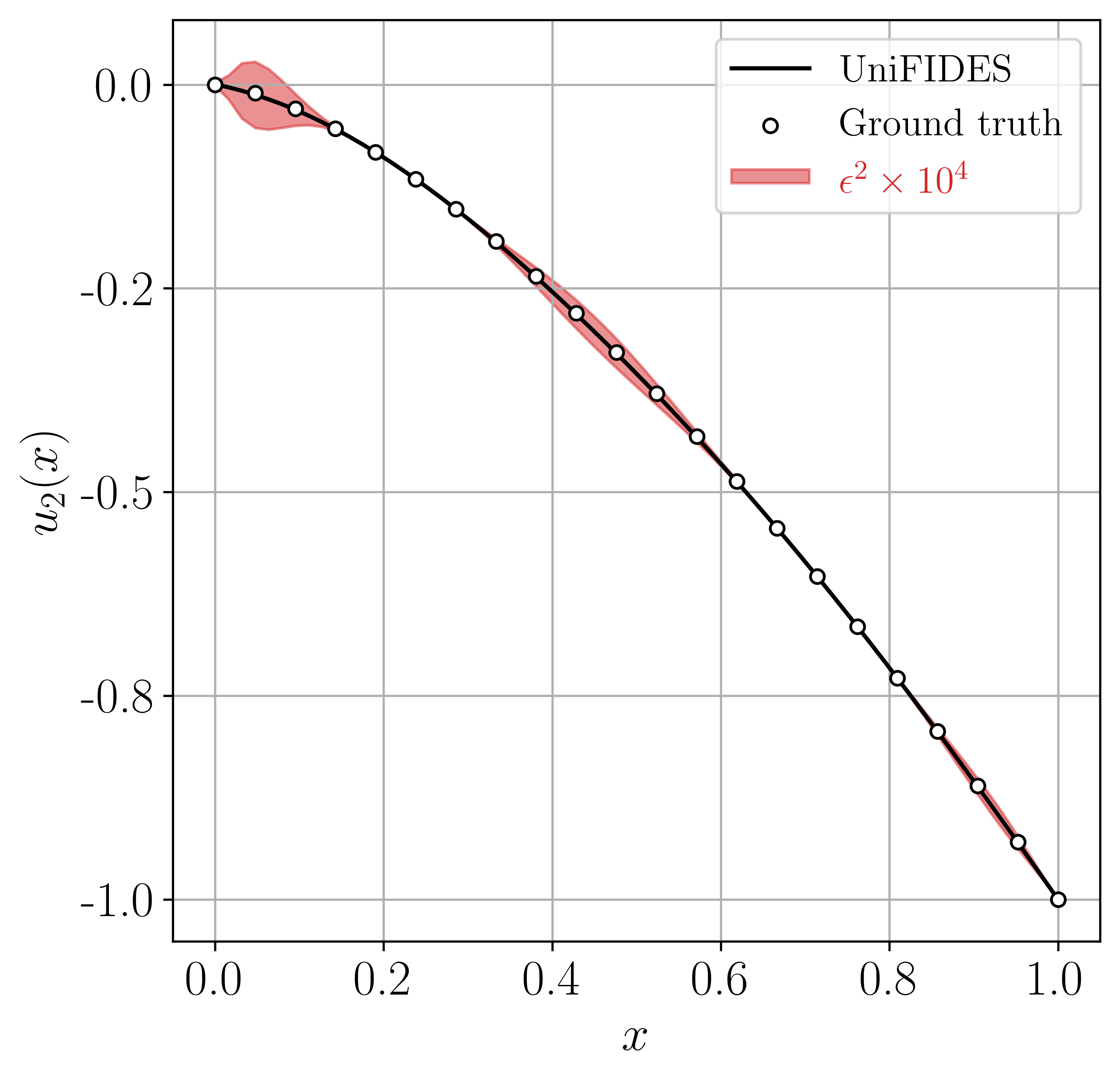}}
        \phantomsubcaption
        \label{fig:Res_C7_b}
    \end{subfigure}
        \caption{The solution by UniFIDES and the exact solution for the system of nonlinear Volterra FIDEs defined in Case 7 and for $\beta=0.5$.}
        \label{fig:Res_C7}
\end{figure}

\subsubsection*{Inverse solution of a system of Volterra FIDEs}
\label{sec:Res_V8}
So far, the IEs and IDEs were assumed known with proper initial and/or boundary conditions. There are, nonetheless, cases in which the exact form or values of these equations are unknown or only partially available, while a [spatio-temporal] measurement of the system is conducted. PINNs have proven to be robust in dealing with such ill-posed problems. To assess the capability of UniFIDES in addressing inverse problems, the same system of FIDEs described in \Cref{eq:V7} is transformed into an inverse problem. The training data is generated using the exact solution for $\beta=0.5$ and $x \in [0,1]$, where $u_1$ and $u_2$ are functions of $x$. In order to replicate real-world scenarios, Gaussian noise with a standard deviation of 0.1 is added to both $u_1$ and $u_2$ vectors. The objective is to recover the integral and derivative orders, denoted as $\alpha$ and $\beta$, respectively. Compared to the exact orders of $\alpha, \beta = (1, 0.5)$, UniFIDES converges to \num{1.025} and \num{0.488}, with exceptional relative errors of \qtylist[list-units = single]{1.025;0.488}{\percent}, respectively. This establishes UniFIDES as a robust solver for both forward and inverse problems across a range of IE and IDE scenarios.

\section{Discussion}\label{sec:Discussion}
This study introduces a resilient and versatile framework for solving a variety of integro-differential equations with both fractional and integer orders, covering generic problems in science and engineering. Despite the inherent difficulties in handling integral equations using PINN-based platforms, our results demonstrate that UniFIDES can tackle such equations in both forward and inverse directions, irrespective of their nonlinearity or operator order. Notably, unlike comparable toolboxes designed for solving FIDEs, which often require case-specific and ad-hoc adjustments, UniFIDES accepts the equations in their original (and continuous) form, allowing for intuitive imposition of boundary or initial conditions.

In all fairness, there exist several limitations associated with UniFIDES in its current form. The first, and probably the most notable limitation, lies in the utilization of a numerical scheme to approximate integrals and fractional derivatives. Similar to other numerical approximations, truncation error is inevitable. The final predictions, however, were comparable in terms of accuracy with, for instance, A-PINN, which obviated the need for discretization. The implemented numerical scheme commits an error of $O(h^2)$, with $h$ being the step size. Although other integral approximations with refined accuracy may reduce this error order, they will nonetheless remain a function of $h$. Moreover, $h$ is assumed constant along each integration (or differentiation) axis.

The hereditary nature of fractional operators and the embedded numerical scheme require the utilization of input vectors in ascending order. This requirement arises from the inherent dependence of fractional operators on the function's historical data, wherein the concept of \textit{history} holds significance only when information is chronologically stored. Therefore, if UniFIDES is tested on non-rectangular geometries, it is vital to pass an ascending vector to \Cref{eq:RL} and evaluate the functional points accordingly. In other words, a fictitious structured mesh is defined and used for fractional operators; other functions may be sampled randomly in the space (or time) but they need to be sorted to ensure accurate pointwise calculations.

In summary, UniFIDES presents enhanced flexibility in both problem definition and implementation, surpassing its current limitations. Future research endeavors should focus on relaxing the constraints of a constant step size and introducing innovative approaches to evaluate spatio-temporal fractional functions. 

\section{Methods}
\label{sec:Methods}

Consider the following 1D Volterra FIDE:
\begin{equation}
\left\{
\begin{aligned}
&\left[ {}^{\beta} \mathcal{D}_x \right] u(x) + \frac{\partial^2 u(x)}{\partial x^2} = f(x) + \left[ {}^\alpha \mathcal{I}_{0}^{x} \right] K(t)u(t)\diff t\\
&x \in \left[ 0, X \right]\\
&u\left(0\right), u\left(X\right)=(u^0, u^X)\\
\end{aligned}
\right.
\label{eq:FIDE_samp}
\end{equation}
where $f(x)$ is a known and discrete function and $K(t)$ is the Abel kernel with a [potential] weak singularity. Such FIDEs are prevalent in anomalous diffusion. The objective in a forward problem is to determine $u(x)$, while in an inverse problem, scattered (and potentially noisy) measurements of $u(x)$ are obtained, with the objective of finding an unknown parameter in the entire problem, e.g., a boundary condition, $f(x)$, $\beta$, or a combination of them. In A-PINN \cite{Yuan2022}, the integral term is replaced with a function $w(x)$. Thus, by taking a derivative of $w(x)$ w.r.t. $x$ for the specific case of $\alpha \in \mathbb{W}$, the above IDE is converted to a system of ODEs with an auxiliary output $w(x)$:

\begin{equation}
\left\{
\begin{aligned}
&\left[ {}^{\beta} \mathcal{D}_x \right] u(x) + \frac{\partial^2 u(x)}{\partial x^2} = f(x) + w(x)\\
&\frac{\partial^\alpha w(x)}{\partial x^\alpha}=K(x)u(x)\\
\end{aligned}
\right.
\label{eq:IDE_Yuan}
\end{equation}
with appropriate boundary conditions. Moreover, A-PINN is restricted to $\beta \in \mathbb{W}$ cases as AD is infeasible for fractional derivatives, too.

$\beta$ and $\alpha$ in \Cref{eq:FIDE_samp}, nonetheless, can be any positive real numbers: $\alpha, \beta \in \mathbb{R_+}$. In UniFIDES, the well-established numerical schemes developed in \cite{Diethelm2005} are used to approximate the fractional operators, which are based on the trapezoidal quadrature rule. The integral of $u(X)$ with a fractional order of $\alpha$ in the RL sense between $[0, X]$ on the grid $x_n=nh : n=0, 1, 2, \dots, N$ where $h=X/N$ is therefore discretized as follows:

\begin{equation}
    \left[ {}^\alpha \mathcal{I}_{0}^{X} \right]u(X) \approx \frac{h^\alpha}{\Gamma(2+\alpha)}\sum_{j=0}^N c_{j,N}(\alpha)u_j
    \label{eq:RL}
\end{equation}
which commits an error of order $O(h^2)$. The quadrature weights $c_{j,N}$ are derived from a product trapezoidal rule:

\begin{equation}
c_{j,N}(\alpha)=
\begin{cases}
    (1+\alpha)N^\alpha - N^{1+\alpha} + (N-1)^{1+\alpha} & \text{if } j = 0 \\
    \left(N-j+1\right)^{1+\alpha}-2\left(N-j\right)^{1+\alpha}+\left(N-j-1\right)^{1+\alpha} & \text{if } 0 < j < N \\
    1 & \text{if } j = N \\
\end{cases}
\label{eq:RL_c}
\end{equation}
To obtain the integral between 0 and $x_n$, $N$ is replaced with $n$ in \Cref{eq:RL,eq:RL_c}. Therefore, for each point $x_n$, these equations are called, which take into account the total memory of past states up to $x_n$.

As mentioned in \Cref{sec:Results}, the $\beta$-th derivative of $u(X)$ w.r.t. $x$ in the RL sense at $X=Nh$ can be written as follows:

\begin{equation}
    \left[ {}^{\beta} \mathcal{D}_x \right] u(X) = \left[ {}^{-\beta} \mathcal{I}_{0}^{X} \right]u(X)
    \label{eq:RLD_Diethelm}
\end{equation}
Again, the fractional derivative at $x_n$ is obtainable by formally replacing $X$ with $x_n$ in \Cref{eq:RLD_Diethelm}. Therefore, fractional derivatives, unlike their integer-order counterparts, are not pointwise operators and depend on the entire functional history. Fractional derivatives in the RL (and also the Caputo) sense using the above scheme are nonetheless definite only for $\beta<1$ as \Cref{eq:RL_c} raises 0 to a negative power for $-\beta=\alpha\leq-1$. Higher-order fractional derivatives are nonetheless obtainable as fractional derivatives, similar to integer-order ones, can be expressed using the chain rule. For instance, the derivative of order $\beta=1.7$ can be decomposed into a first-order derivative and a 0.7-th-order fractional derivative.

In summary, \Cref{eq:RL,eq:RL_c} are valid for $\alpha >-1$, which allows us to calculate fractional derivatives of orders $0\leq \beta<1$ and any fractional (or integer-order) integrals. For derivatives in each dimension, one boundary (or initial) condition is needed. For instance, \Cref{eq:FIDE_samp} requires two boundary values.

The neural networks are initially sub-classed from \texttt{TensorFlow.keras.Model} and modified to handle fractional operations. The second-order derivative is obtainable through \texttt{GradientTape}, which is a context manager that records operations for AD in \texttt{TensorFlow}. Finally, the fractional terms in \Cref{eq:FIDE_samp} are directly calculated using \Cref{eq:RL,eq:RL_c}. The forward problem defined in \Cref{eq:FIDE_samp}, therefore, is converted to a loss minimization task:

\begin{equation}
    \phi=\phi_\mathrm{Res}+\phi_\mathrm{BC}
    \label{eq:Loss}
\end{equation}
where $\phi$ is the compound loss function and $\phi_\mathrm{Res}$ is the equation residual:

\begin{equation}
    \phi_\mathrm{Res}=\frac{1}{N}\sum_{n=0}^{N} \left\{\left[ {}^{-\beta} \mathcal{I}_{0}^{x_n} \right]u_p(x_n) + \frac{\partial^2 u_p(x_n)}{\partial x_n^2} -\left( f(x_n) + \left[ {}^\alpha \mathcal{I}_{t=0}^{t=x_n} \right] K(t)u_p(t)\diff t\right)\right\}^2
    \label{eq:Loss_eq}
\end{equation}
where $u_p$ is the UniFIDES prediction of the output. The boundary condition loss ($\phi_{BC}$) is calculated as the MSE of the NN prediction at the boundaries and the ground-truth information on the boundaries. For this particular example, Dirichlet BCs are imposed as followes:

\begin{equation}
    \phi_{BC} = \textrm{MSE}(u_{p}^0, u^{0})+\textrm{MSE}(u_{p}^X, u^{X})=
    \frac{1}{N_b}\sum_{m=1}^{N_b} (u_{p,m}^0 - u^0)^2+
    \frac{1}{N_b}\sum_{m=1}^{N_b} (u_{p,m}^X - u^X)^2
    \label{eq:Loss_bc}
\end{equation}

In each iteration, the compound loss is computed, and the \texttt{Adam} optimizer from the \texttt{tf.keras.optimizers} module is utilized to update the model parameters, i.e., $\theta$ in \Cref{fig:Arch}. For more details on the UniFIDES architecture, training process, and hyperparameters, the readers are referred to \Cref{sec:SI-train}.

\section*{Code availability}\label{sec:Code}

All the source codes to reproduce the results in this study are available in the \href{https://github.com/procf/RhINNs}{GitHub repository}.

\section*{Declarations}

Authors are thankful for insightful discussions with Dr. Deepak Mangal, and
also acknowledge the support from the National Science Foundation’s DMREF
program through Award \#2118962.

\newpage

\setcounter{section}{0}
\setcounter{equation}{0}
\setcounter{figure}{0}
\setcounter{table}{0}
\setcounter{page}{1}

\renewcommand{\thepage}{S\arabic{page}}
\renewcommand{\thesection}{S\arabic{section}}
\renewcommand{\theequation}{S\arabic{equation}}
\renewcommand{\thetable}{S\arabic{table}}
\renewcommand{\thefigure}{S\arabic{figure}}
\renewcommand{\thealgocf}{S\arabic{algocf}}

\section{UniFIDES training process and hyperparameter tuning}\label{sec:SI-train}
Similar to other numerical solvers for forward problems, two sets of data are needed in UniFIDES: 1. boundary (or initial) condition points ($N_b$ points in \Cref{eq:Loss_bc}), and 2. residual (collocation) points ($N$ points in \Cref{eq:Loss_eq}). The former can be randomly distributed across the boundaries.\footnote{This is valid only for non-fractional (Dirichlet, integer-order Neumann, etc.) BCs. If a BC is subject to a fractional operator, the BC points need to be strictly ascending.} For residual calculations of fractional terms, and as discussed in \Cref{sec:Discussion}, the points have to increase monotonically as fractional operators depend on \textit{previous} information. These vectors follow the problem dimensions, i.e., for a 2D problem on $x$ and $y$, two 1D vectors are defined and concatenated. Integer-order terms may be queried on randomly-generated vectors.

After defining these vectors, a feed-forward, fully connected neural network is instantiated from \texttt{tf.keras.Model} with 3 hidden layers, each containing 16 neurons (see below for sensitivity analysis). Weights and biases are initialized using the \verb|glorot_normal| method. Also, the widely-used \texttt{tanh} activation function is selected for hidden layers as a robust and differentiable function. The last (output) layer has a linear activation function.

After instantiating the NN, the solver is defined. The fractional integral in the Riemann-Liouville sense is implemented following the algorithm described in \Cref{eq:RL,eq:RL_c}. As discussed in \Cref{sec:Methods}, fractional derivatives are accessible by simply calling the same function but with a negative order. The integer-order derivatives are obtainable using \texttt{TensorFlow}'s built-in \texttt{GradientTape} class.

After obtaining all the required derivatives and integrals, the compound loss ($\phi$) is calculated by imposing the equation residual loss and also the BC loss; see \Cref{eq:Loss,eq:Loss_eq,eq:Loss_bc}. The gradient of $\phi$ w.r.t. the trainable variables ($\theta$) is then calculated using \texttt{GradientTape}. Finally, the \texttt{Adam} optimizer with a piecewise learning rate adjusts $\theta$ to satisfy the constraints imposed by equation and BC losses. Although there is no guarantee that the achieved convergence is a global minimum, the loss order (and the model prediction) nonetheless gives the user a hint as to whether the achieved convergence is close enough to this global minimum or not.

The training is halted either by reaching the maximum number of allowed iterations ($N_\mathrm{it}$) or if the loss is plateaued. The latter is achieved by monitoring the last 20 loss values. If the loss in the last 20 iterations does not change below a threshold (e.g., \num{1e-3}), the achieved convergence is assumed to have plateaued. The loss history ($\phi$) as a function of the iteration number ($N_\mathrm{it}$) for Cases 1 and 3 (Fredholm and Volterra IDEs, respectively) is also presented in \Cref{fig:SI-Hist}. To tackle an inverse problem, a trainable vector containing the unknown parameters is defined. Therefore, the trainable variables in this case are the weights, biases, and the unknown parameter(s). The training algorithm for an example 2D forward problem is summarized in Algorithm \ref{alg:UniFIDES}. All training tasks are performed on a MacBook Pro (M1 Max, 64 GB RAM) without GPU acceleration.

\begin{figure}
    \centering
    \begin{subfigure}[b]{0.46\textwidth}
        \stackinset{l}{0.2in}{t}{-.2in}{(a)}{\includegraphics[width=\textwidth]{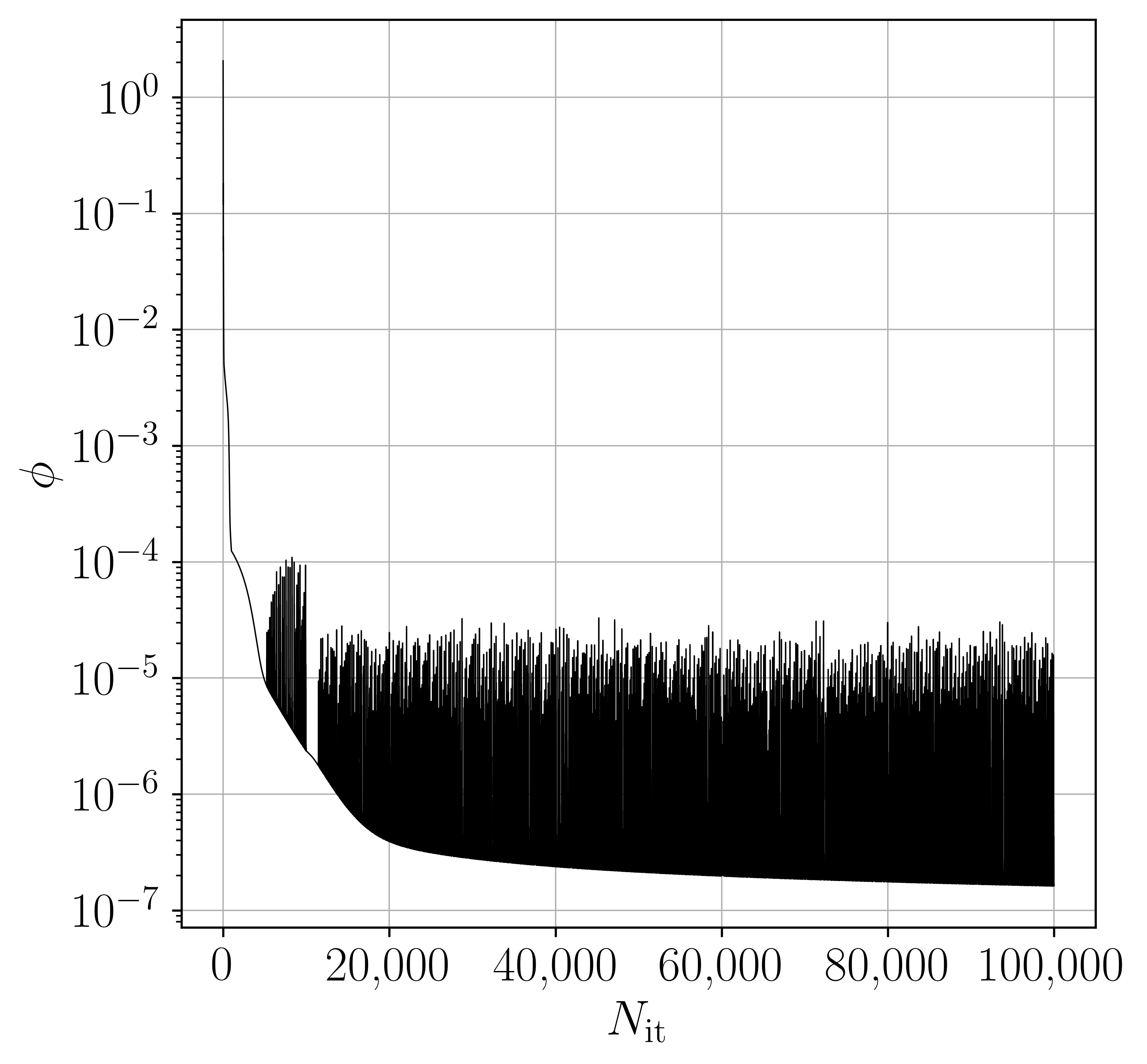}}
        \phantomsubcaption
        \label{fig:SI-Hist_a}
    \end{subfigure}
    \hfill 
    \begin{subfigure}[b]{0.45\textwidth}
        \stackinset{l}{0.2in}{t}{-.2in}{(b)}{\includegraphics[width=\textwidth]{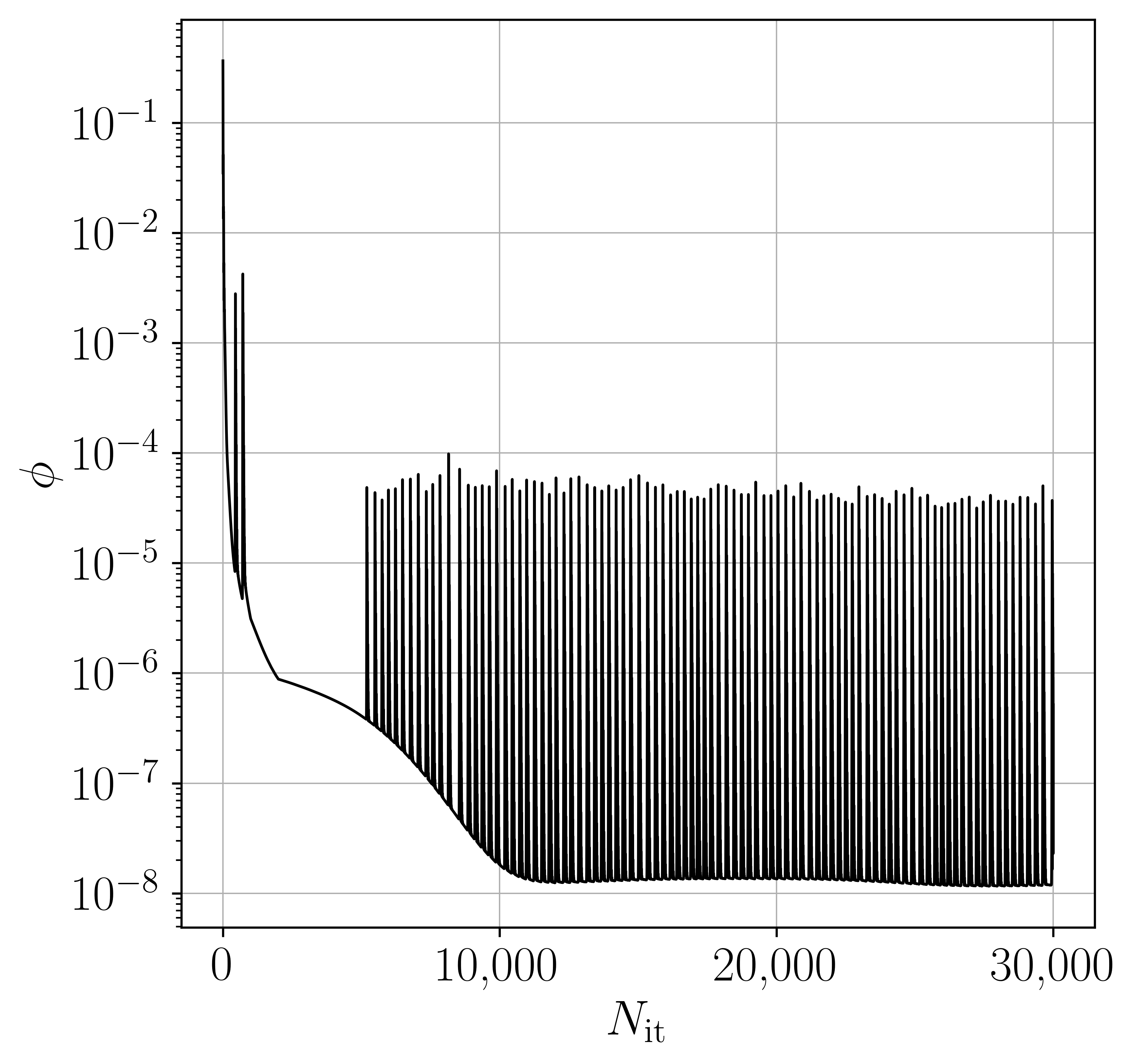}}
        \phantomsubcaption
        \label{fig:SI-Hist_b}
    \end{subfigure}
        \caption{The loss history ($\phi$) as a function of the number of UniFIDES iterations ($N_\mathrm{it}$) for (a) 1D Fredholm IDE (Case 1) and (b) 1D Volterra IDE (Case 3). Other cases follow a similar loss plateau pattern.}
        \label{fig:SI-Hist}
\end{figure}

\begin{algorithm}[h!]
    \SetAlgoLined
    \KwData{Input data: $N$ (no. of residual points), $N_b$ (no. of boundary condition points), $N_\mathrm{it}$ (no. of iterations)}
    \KwResult{$u(x,y)$}
    Initialize residual points ($X_r$) using $x_{\text{\normalfont{min}}}$, $x_{\text{\normalfont{max}}}$, $y_{\text{\normalfont{min}}}$, $y_{\text{\normalfont{max}}}$,
    $N_r$;\
    
    Initialize boundary condition points ($X_b$) using 
    $x_{\text{\normalfont{min}}}$, $x_{\text{\normalfont{max}}}$, $y_{\text{\normalfont{min}}}$, $y_{\text{\normalfont{max}}}$,
    $N_b$;\
    
    Instantiate an NN using the given hyperparameters\;
    \For{$i \leftarrow 1$ \KwTo $N_{it}$}{
        Get $u_p$ and all its required integrals/derivatives over $X_r$\;
        Calculate the residual of the given equation over $X_r$ and obtain its MSE ($\phi_\mathrm{Res}$)\;
        Get $u_p^0$ at BC(s) and calculate the MSE of $u_p^0$ against the known BC(s) ($\phi_{BC}$)\;
        Construct $\phi$ using a given error heuristic (e.g., \Cref{eq:Loss})\;
        Get the gradient of $\phi$ w.r.t. the trainable variables (NN weights and biases)\;
        Apply the gradients to the trainable variables to minimize $\phi$ using Adam optimization with a learning rate schedule.
    }
    \caption{Forward UniFIDES algorithm to solve $u(x,y)$}\label{alg:UniFIDES}
\end{algorithm}

One crucial aspect of NN-based solvers is the role of their hyperparameters, i.e., the number of hidden layers ($n_l$), neurons per layer ($n_n$), etc. The effect of $n_l$ and $n_n$ is summarized in \Cref{tab:SI-n_l,tab:SI-n_n}, respectively. As expected, the execution time increases as $n_l$ or $n_n$ increase. However, and for this particular case, having 3 layers and 16 neurons per layer (highlighted in both tables) results in the lowest MSE, compared to other values. These values are therefore selected throughout this work to construct the NN architecture (Except for Case 1, which employed 2 layers and 20 neurons per layer, in accordance with \cite{Yuan2022}). It is crucial to mention that these hyperparameters do not necessarily yield the same MSE and execution time for other cases covered in this work. The selected hyperparameters nonetheless provide a reasonable compromise between the solution accuracy and execution time for other cases as well.

\begin{table}
\centering
\caption{The effect of the number of layers ($n_l$) on the prediction MSE against the exact solution and execution time ($t$) for the 1D Volterra FIE (Case 5 in \Cref{sec:Results}). The number of collocation (residual) points ($N$) and neurons per layer ($n_n$) is set to 64 and 16, respectively. The maximum number of iterations ($N_\mathrm{it}$) is set to \num{3e4}.}\label{tab:SI-n_l}
\begin{tabularx}{\textwidth}{Y Y Y}
\hline\hline
$n_l$ & MSE & $t$ [\unit{\minute}] \\ \hline
2        & \num{6.55e-7}   & \num{2.85}          \\
\rowcolor{red!15} 
3        & \num{3.72e-7}   & \num{3.20}          \\ 
4        & \num{4.61e-7}   & \num{3.52}          \\ 
5        & \num{8.72e-7}   & \num{3.67}          \\ 
6        & \num{6.27e-7}   & \num{3.88}          \\ \hline\hline
\end{tabularx}
\end{table}

Another important hyperparameter is the activation function for hidden layers. In this work, the \texttt{tanh} activation function is employed, as frequently used by the PINN community due to its differentiability. This activation function nonetheless may result in \textit{vanishing gradients}, as for small or large input values, the output signal of \texttt{tanh} is plateaued. However, compared to other popular activation functions such as \texttt{sigmoid} and \texttt{ReLU}, \texttt{tanh} remained the best option for all cases tested here. Moreover, the effect of kernel initialization was insignificant and this hyperparameter was set to \verb|glorot_normal|, as mentioned above.

\begin{table}[h!]
\centering
\caption{The effect of the number of neurons ($n_n$) on the prediction MSE against the exact solution and execution time ($t$) for the 1D Volterra FIE (Case 5 in \Cref{sec:Results}). The number of layers ($n_l$) and collocation points ($N$) is set to 3 and 64, respectively. The maximum number of iterations ($N_\mathrm{it}$) is set to \num{3e4}.}
\label{tab:SI-n_n}
\begin{tabularx}{\textwidth}{Y Y Y}
\hline\hline
$n_n$ & MSE & $t$ [\unit{\minute}] \\ \hline
8        & \num{4.71e-7}   & \num{3.17}          \\ 
\rowcolor{red!15} 
16        & \num{3.72e-7}   & \num{3.20}          \\ 
32        & \num{6.10e-7}   & \num{3.60}          \\ 
64        & \num{9.39e-7}   & \num{4.54}          \\ \hline\hline
\end{tabularx}%
\end{table}

Moreover, as UniFIDES adopts a numerical scheme to approximate fractional operators, the number of collocation (residual) points ($N$) also plays an important part by altering the step size, $h$, in \Cref{eq:RL}. \Cref{fig:SI-N} shows the effect of $N$ on the prediction MSE against the exact solution ($u_e$) and also the execution time ($t_{\mathrm{Exec.}}$) for Case 5. For this particular case, having 64 collocation points appears to be an acceptable compromise between the MSE and $t_{\mathrm{Exec.}}$. The overall prediction, nonetheless, closely mimics the exact solution once the MSE drops below $\num{1e-6}$. In practice, the number of collocation points was varied for each case to achieve accurate results while keeping $t_{\mathrm{Exec.}}$ checked; see \Cref{tab:MSE} for the values of $N$ for each case.

\begin{figure}[h!]
  \centering
  \includegraphics[width=.7\linewidth]{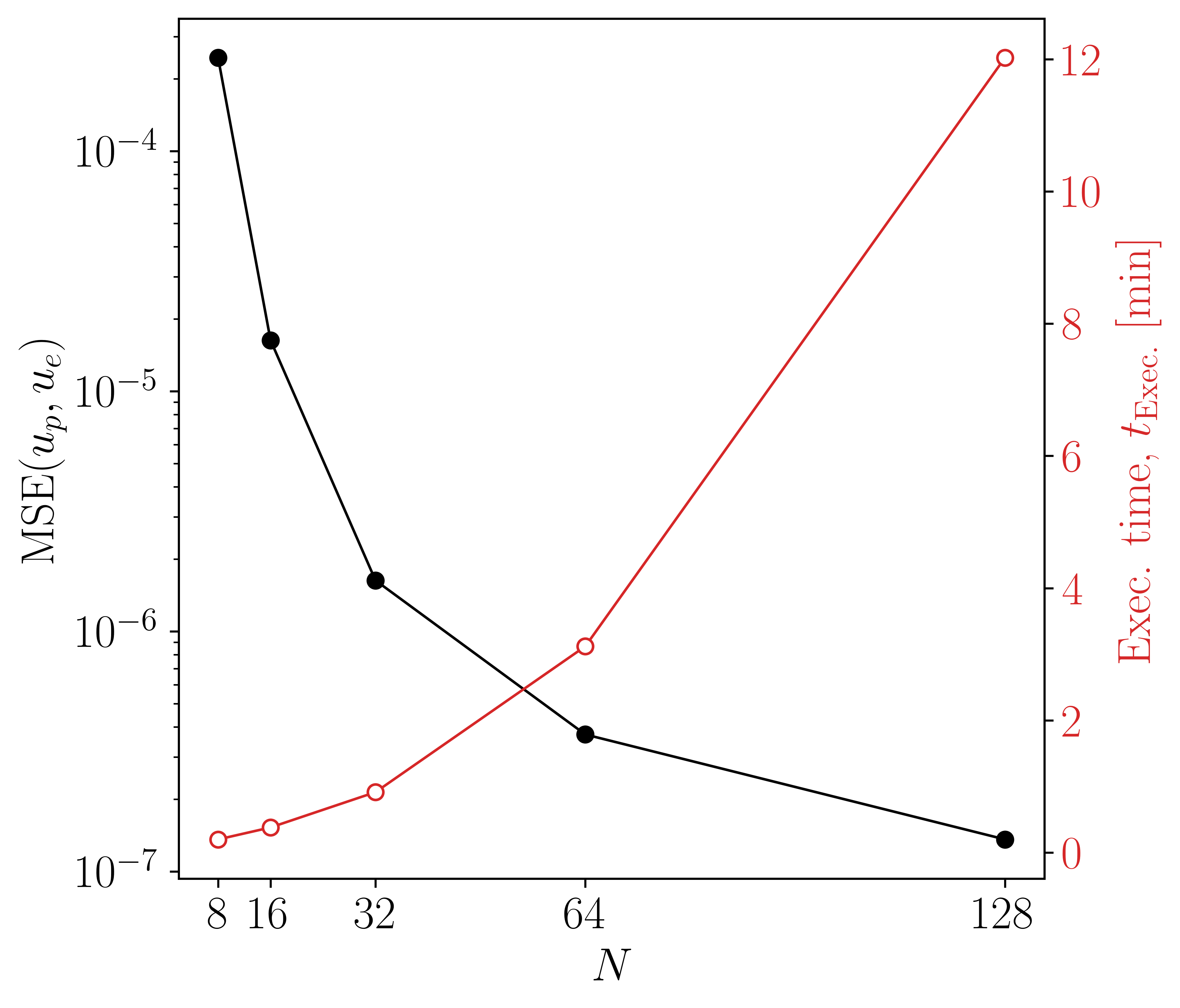}
  \caption{The effect of the number of collocation (residual) points on the prediction MSE against the exact solution (left axis) and the execution time (right axis in red) for the 1D Volterra FIE (Case 5). Here, $n_l$ and $n_n$ are set to 3 and 16, respectively. The maximum number of iterations ($N_\mathrm{it}$) is set to \num{3e4}.}
  \label{fig:SI-N}
\end{figure}

\section{A-PINN benchmarking procedure}
\label{sec:SI-APINN}
In this section, the A-PINN methodology is described using Case 4, which is an integer-order integral equation. In A-PINN, the integral term is parametrized as $v(x,y)$:
\begin{equation}
\left\{
\begin{aligned}
&u(x,y)=f(x,y)+v(x,y) \\
&f(x,y)=x\sin(y)\left(1-\frac{x^2\sin^2{y}}{9}\right)+\frac{x^6}{10}\left(\frac{\sin(2y)}{2}-y\right)\\
& v(x,y)= \left[ {}^1 \mathcal{I}_{0}^{y} \right]\left[ {}^1 \mathcal{I}_{0}^{x} \right] \left(xt^2+\cos(s)\right)u^2(t,s) \diff t \diff s\\
&(x, y) \in \left[ 0, 0.5 \right], \left[ 0, 1 \right]\\
&u\left(0, 0\right)=0
\end{aligned}
\right.
\label{eq:SI_4-APINN}
\end{equation}
Using Leibniz' integral rule, the partial derivative of $v(x,y)$ w.r.t. the outer integral variable ($y$) is accessible, assuming a continuous partial derivative of the integrand, among other assmuptions:
\begin{equation}
    \frac{\partial v(x,y)}{\partial y}=\int_{0}^{x} \left[ xt^2+\cos(y)\right]u^2(t,y)\diff{t}=w(x,y)
    \label{eq:SI-dvdy}
\end{equation}
The Leibniz's rule is applied once again by taking a partial derivative of $w(x,y)$ w.r.t. $x$:
\begin{equation}
\begin{aligned}
    \frac{\partial w(x,y)}{\partial x}=\left[x^3+\cos(y)\right]u^2(x,y) + \int_{0}^{x} t^2u^2(t,y)\diff{t} = \\\left[x^3+\cos(y)\right]u^2(x,y) + p(x,y)
    \label{eq:SI-dvdx}
    \end{aligned}
\end{equation}
Finally, the last integral may be eliminated by taking another partial derivative of $p(x,y)$ w.r.t. $x$:

\begin{equation}
    \frac{\partial p(x,y)}{\partial y}= x^2u^2(x,y)
    \label{eq:SI-dpdx}
\end{equation}

In summary, three auxiliary outputs ($v(x,y)$, $w(x,y)$, and $p(x,y)$) are defined, and this case is implemented as follows using A-PINN:

\begin{equation}
\left\{
\begin{aligned}
&u(x,y)=f(x,y)+v(x,y) \\
&\frac{\partial v(x,y)}{\partial y} = w(x,y)\\
&\frac{\partial w(x,y)}{\partial x} = \left[x^3+\cos(y)\right]u^2(x,y + p(x,y))\\
&\frac{\partial p(x,y)}{\partial x}=x^2u^2(x,y)\\
&f(x,y)=x\sin(y)\left(1-\frac{x^2\sin^2{y}}{9}\right)+\frac{x^6}{10}\left(\frac{\sin(2y)}{2}-y\right)\\
&(x, y) \in \left[ 0, 0.5 \right], \left[ 0, 1 \right]\\
&\left\{u, v, w, p\right\}\left(0, 0\right)=0
\end{aligned}
\right.
\label{eq:SI_4-APINN-summary}
\end{equation}
As can be seen, this problem redefinition cleverly eliminates the need for integration by converting the integral equation into a system of algebraic and ordinary differential equations. The initial conditions for \( v(x,y) \), \( w(x,y) \), and \( p(x,y) \), however, are established by explicitly inserting the exact solution into \Cref{eq:SI_4-APINN-summary} and examining the nested variables. Also, another significant drawback of A-PINN is the assumption of ``integral elimination.'' For example, the integral in the equation \( u(x) = \int_{0}^{x} \sin(tx) \, dt \) cannot be eliminated using Leibniz's integral rule. However, a simple substitution, \( p = tx \), yields the exact solution \( u(x) = \frac{1 - \cos(x^2)}{x} \). This problem can be readily implemented in UniFIDES in the original form, and despite the singularity at \( x = 0 \), it avoids the need for ad-hoc modifications that A-PINN would require.

In \Cref{tab:MSE}, the UniFIDES' predictions are benchmarked against those of A-PINN. It is worth mentioning that A-PINN can only tackle integer-order cases, as the Leibniz's rule eliminates only integer-order integral terms.

\section{On the difference between Fredholm and Volterra integrals}
\label{sec:SI-FV}
Fredholm integrals have definite bounds. In other words, they are equivalent to a scalar and not a vector. The fractional heuristic \cite{Diethelm2005} adopted in this work, however, essentially looks back into the history of the function being integrated. For instance, consider the below Fredholm integral:

\begin{equation}
    u(x) = \int_{0}^{1} K(x) u(t) \, dt=K(x)\int_{0}^{1} u(t) \, dt
    \label{eq:SI-Fredholm}
\end{equation}
In this instance, the kernel $K(x)$ is independent of $t$, permitting it to be factored out of the integral expression. Therefore, the integral itself is a scalar. \Cref{eq:RL,eq:RL_c}, on the other hand, return the integral from zero up to each point $x_n$ between 0 and 1. To speed up the computations, the numerical scheme for Fredholm integrals is called only once and only for $x_n=1$. This method efficiently calculates the definite integrals. However, for Volterra integrals, the integral for each point needs to be obtained based on prior information. Therefore, a \texttt{for} loop is used to iterate through all points in each dimension and store the vectorized integrals. The implementation of both integral types can be found \href{https://github.com/procf/RhINNs}{here}.

\section{Partial fractional derivative calculation}\label{sec:SI-Deriv}
The \texttt{GradientTape} \href{https://www.tensorflow.org/api_docs/python/tf/GradientTape}{context manager} on \texttt{TensorFlow} automatically calculates the partial derivative. For instance, running the following code on \texttt{TensorFlow} v2:
\begin{mintedbox}{python}
import tensorflow as tf

x = tf.Variable(2.0, dtype=tf.float32)
y = tf.Variable(3.0, dtype=tf.float32)

u = x**2 + 2*y

with tf.GradientTape() as tape:
    tape.watch(x)
    u = x**2 + 2*y
tape.gradient(u, x)
\end{mintedbox}
\noindent yields \texttt{<tf.Tensor: shape=(), dtype=float32, numpy=4.0>}.

However, for multi-dimensional problems with the numerical scheme introduced in \Cref{eq:RL,eq:RL_c}, extra caution is needed when calculating partial derivatives. Among methods to handle partial derivatives numerically, we employed a rather straightforward workaround by simply reshaping the NN prediction into a grid and performing the fractional operations along the axis for which the derivative is required. For instance, consider the following code to handle the derivative of the model prediction, \texttt{u}, w.r.t. \texttt{t}:

\begin{mintedbox}{python}
ud = tf.reshape(u, [Nx, Nt])
ufd = tf.TensorArray(tf.float32, size=Nx)
for i in tf.range(Nx):
    ufd = ufd.write(i, [self.RL(beta, ud[i, :], index, h_t) for index in range(Nt)])
ufd = tf.reshape(ufd.stack(), (-1, 1))
\end{mintedbox}

In the following code snippet, the points \texttt{Nx} and \texttt{Nt} are distributed linearly, and the input vector is constructed using \texttt{tf.meshgrid}, with subsequent flattening of the meshes into a 2D vector. The method \texttt{RL} is applied to each row of \texttt{ud}, representing the reshaped model prediction. This methodology can be extended to address higher-dimensional problems by keeping all variables constant except for the one for which the derivative is required. This method, nonetheless, is arguably not the most efficient one, and further optimization is warranted to improve UniFIDES' computation efficiency. 

\section{Supplementary results}\label{sec:SI-Res}
In this section, the UniFIDES' predictions for three additional Fredholm IEs are presented to complement the findings presented in \Cref{sec:Results}.

The first supplementary case (Case S1) is a 1D Fredholm IE \cite{Avazzadeh2011}:

\begin{equation}
\left\{
\begin{aligned}
&u(x) = \tan{(x)} - \left[ {}^1 \mathcal{I}_{-\pi/3}^{\pi/3} \right] e^{\arctan(x)}u(t) \, \mathrm{d}t \\
&x \in \left[ -1, 1 \right]\\
&u\left( -1 \right)\approx 1.56
\end{aligned}
\right.
\label{eq:SF1}
\end{equation}
with an exact solution of $u(x)=\tan{(x)}$. This integral signifies that the accumulative behavior of the integral kernel between $-\pi/3$ and $\pi/3$ affects the $u(x)$ distribution. The solution for this case is plotted in \Cref{fig:Res_S1_2_a} with an overall MSE of \num{2.23e-6}.

\begin{figure}
    \centering
    \begin{subfigure}[b]{0.46\textwidth}
        \stackinset{l}{0.2in}{t}{-.2in}{(a)}{\includegraphics[width=\textwidth]{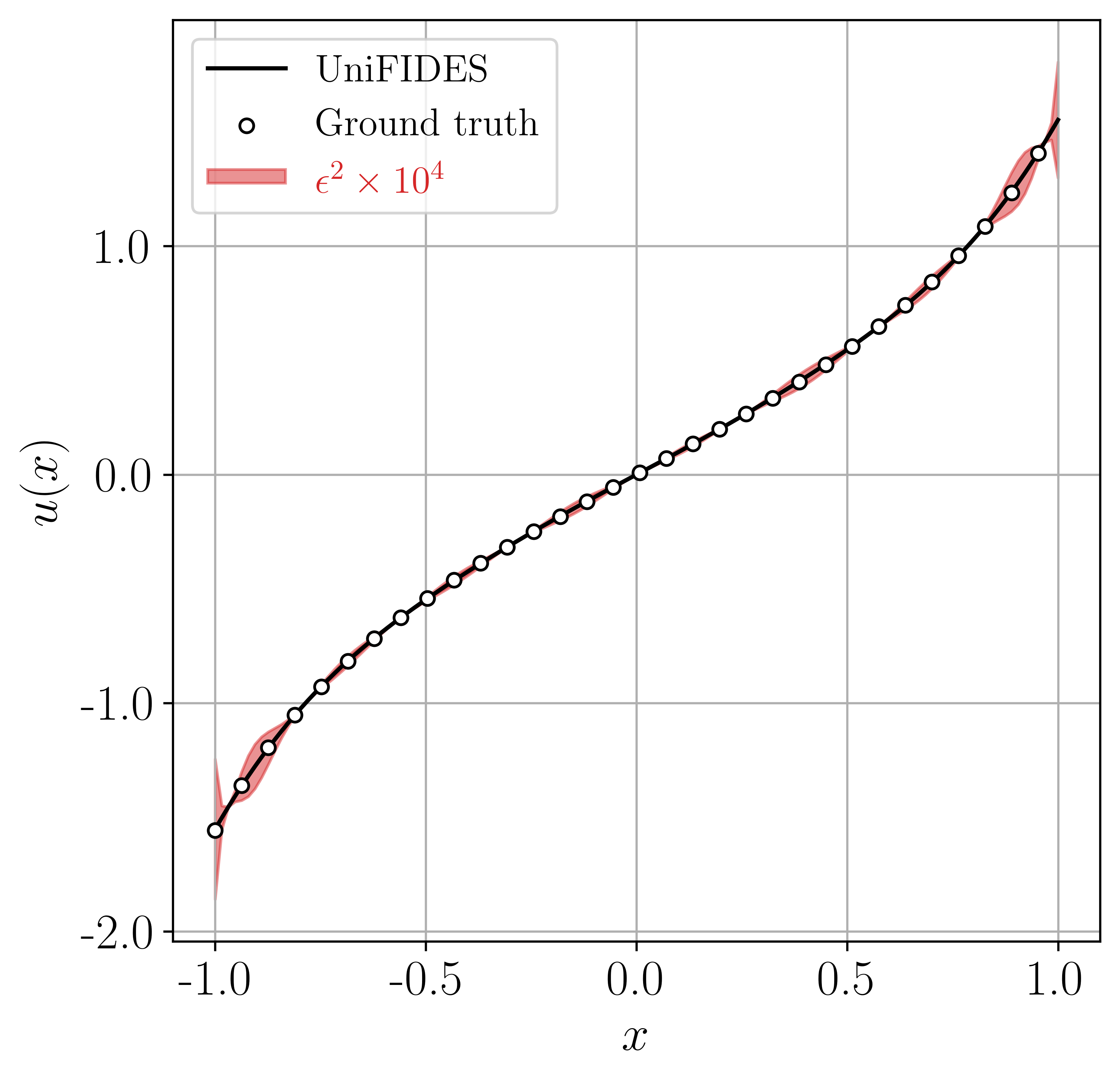}}
        \phantomsubcaption
        \label{fig:Res_S1_2_a}
    \end{subfigure}
    \hfill 
    \begin{subfigure}[b]{0.47\textwidth}
        \stackinset{l}{0.2in}{t}{-.2in}{(b)}{\includegraphics[width=\textwidth]{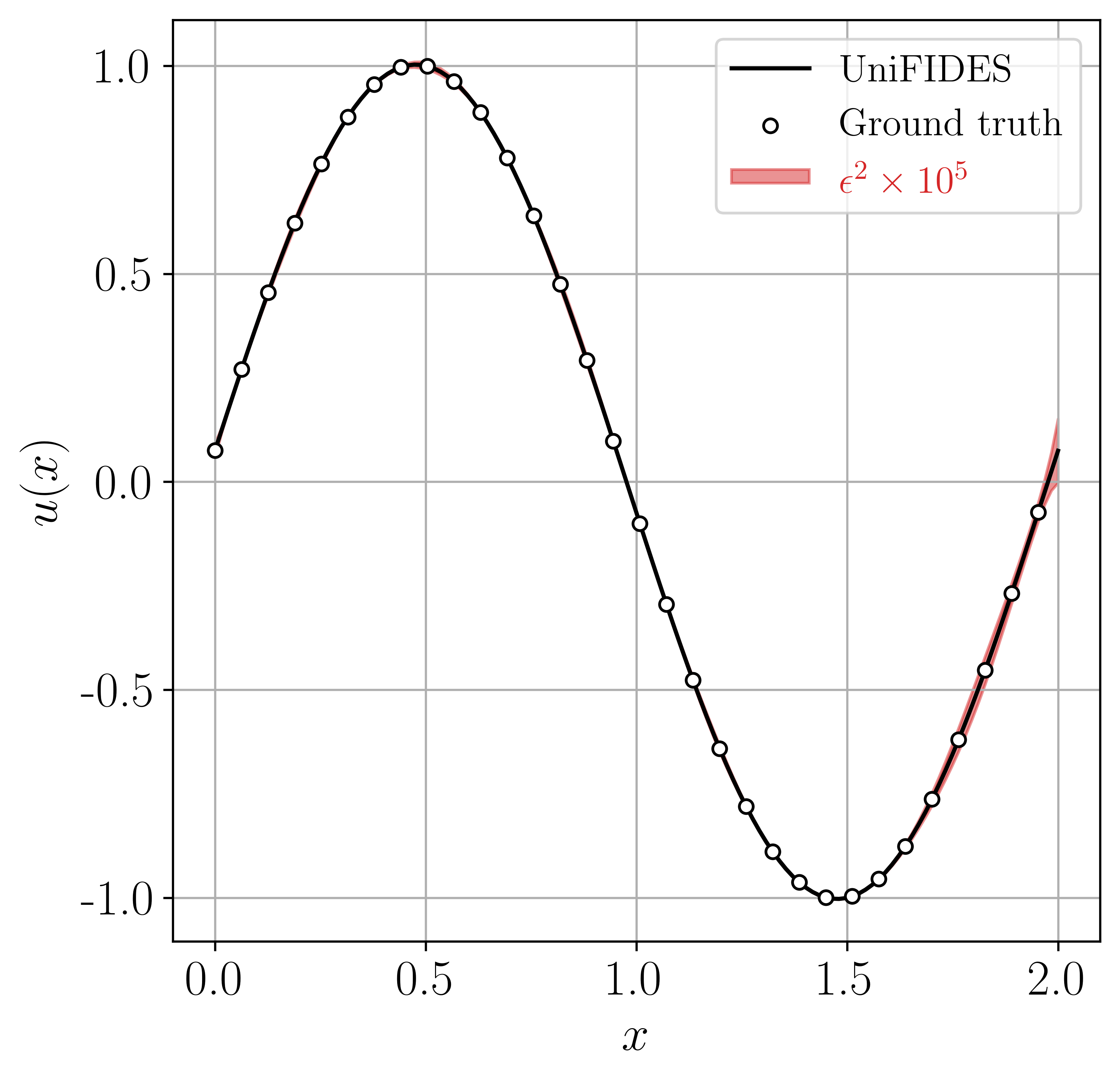}}
        \phantomsubcaption
        \label{fig:Res_S1_2_b}
    \end{subfigure}
        \caption{The solution by UniFIDES and the exact solution for (a) Case S1 and (b) Case S2. For both cases, $N$ is set to 128 with the default NN hyperparameters introduced in \Cref{sec:SI-train}. These Fredholm IEs demonstrate UniFIDES' robustness in handling nonzero ICs.}
        \label{fig:Res_S1_2}
\end{figure}

Case S2 is another nonlinear 1D Fredholm IE \cite{Aziz2013}:

\begin{equation}
\left\{
\begin{aligned}
&u(x) = \sin{(\pi x)} +\frac{1}{5} \left[ {}^1 \mathcal{I}_{0}^{1} \right] \cos{(\pi x)}\sin{(\pi t)} u^3(t) \, \mathrm{d}t \\
&x \in \left[ 0, 2 \right]\\
&u\left( 0 \right)\approx 0.075
\end{aligned}
\right.
\label{eq:SF2}
\end{equation}
with an exact solution of $u(x)=\sin{(\pi x)}+\frac{20-\sqrt{391}}{3}\cos{(\pi x)}$. The solution of this case is shown in \Cref{fig:Res_S1_2_b}, which attained an MSE of \num{8.19e-8}.

The final supplementary case (Case S3) is the same as \Cref{eq:F2} but for $(x,y,z) \in \left[0, 2\right]$ \cite{Mahdy2023}. Therefore, the solution peaks at 64 at the maximum input values; see \Cref{fig:Res_S3}.

\begin{figure}
  \centering
  \includegraphics[width=\linewidth]{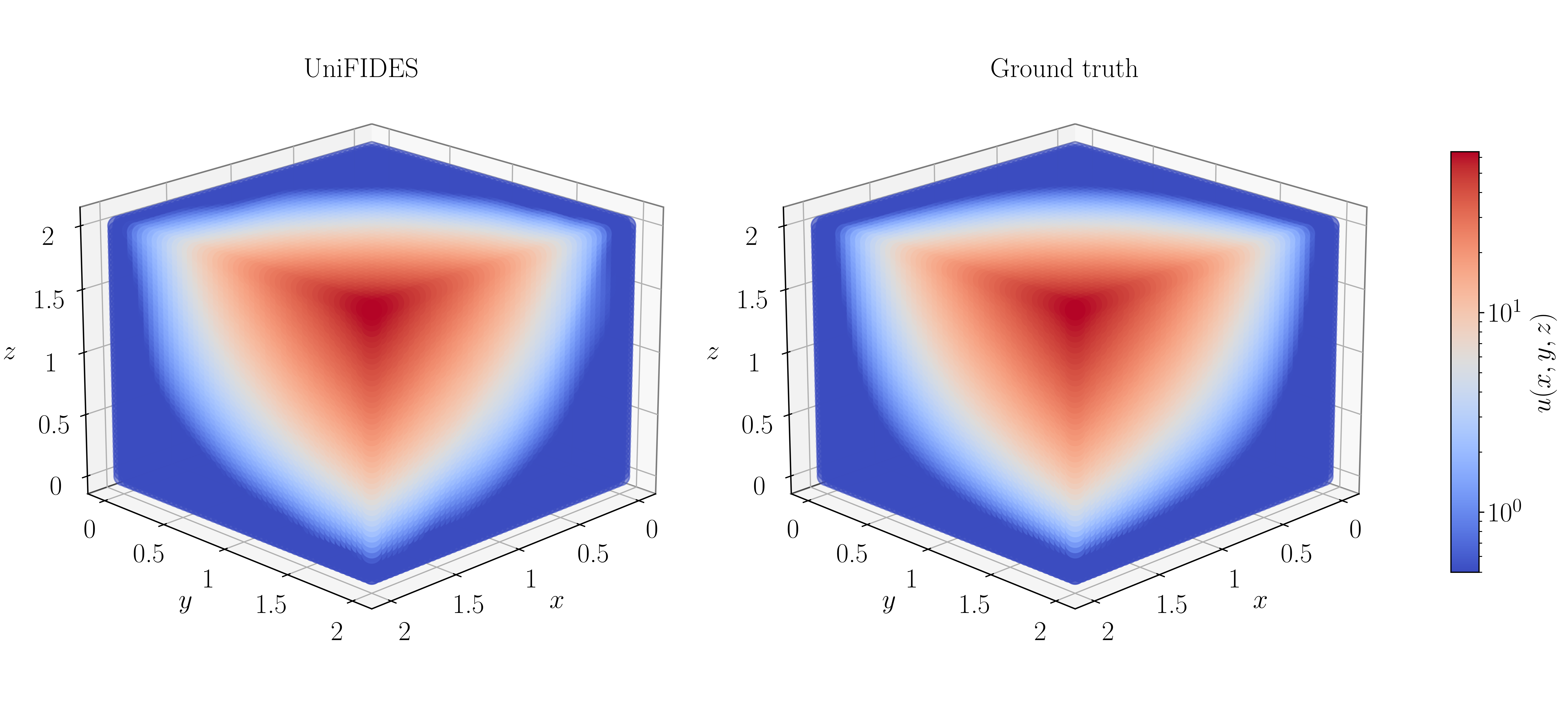}
  \caption{The solution by UniFIDES and the exact solution for the 3D Fredholm IE (Case S3), which follows the same form as in \Cref{eq:F2} but for $(x,y,z) \in \left[0, 2\right]$. The prediction MSE is \num{7.85e-3}, which is acceptable given the solution range.}
  \label{fig:Res_S3}
\end{figure}

\clearpage

\bibliography{library}

\end{document}